%% file: Main.tex
\documentclass[sigconf,nonacm]{acmart}

\usepackage{soul}
\usepackage{enumitem}
\usepackage{amsmath}
\usepackage{algpseudocode}
\usepackage{algorithm}
\usepackage{hyperref}
\usepackage{graphicx}

\usepackage{geometry}
\usepackage{array}
\usepackage{multirow}
\usepackage{multicol}
\usepackage{colortbl}
\usepackage{subfig}
\usepackage{caption}
\usepackage{amsmath}
\usepackage{makecell}
\usepackage{booktabs}

\usepackage{subcaption}

\newcolumntype{R}[1]{>{\raggedleft\arraybackslash}p{#1}}

\begin{document}
\setlength{\abovedisplayskip}{3pt}
\setlength{\belowdisplayskip}{3pt}

\newcolumntype{C}[1]{>{\centering\arraybackslash}p{#1}}
\title{SurvAttack: Black-Box  Attack On Survival Models \\ through Ontology-Informed EHR Perturbation}

\author{Mohsen Nayebi Kerdabadi}
\email{mohsen.nayebi@ku.edu}
\affiliation{%
  \institution{University of Kansas}
  \city{Lawrence}
  \state{KS}
  \country{USA}
}

\author{Arya Hadizadeh Moghaddam}
\email{a.hadizadehm@ku.edu}
\affiliation{%
  \institution{University of Kansas}
  \city{Lawrence}
  \state{KS}
  \country{USA}
}

\author{Bin Liu}
\email{bin.liu1@mail.wvu.edu}
\affiliation{%
  \institution{West Virginia University}
  \city{Morgantown}
  \state{WV}
  \country{USA}
}

\author{Mei Liu}
\email{mei.liu@ufl.edu}
\affiliation{%
  \institution{University of Florida}
  \city{Gainesville}
  \state{FL}
  \country{USA}
}

\author{Zijun Yao}
\email{zyao@ku.edu}
\authornote{Corresponding author.}
\affiliation{%
  \institution{University of Kansas}
  \city{Lawrence}
  \state{KS}
  \country{USA}
}





\begin{abstract}
Survival analysis (SA) models have been widely studied in mining electronic health records (EHRs), particularly in forecasting the risk of critical conditions for prioritizing high-risk patients. However, their vulnerability to adversarial attacks is much less explored in the literature. Developing black-box perturbation algorithms and evaluating their impact on state-of-the-art survival models brings two benefits to medical applications. First, it can effectively evaluate the robustness of models in pre-deployment testing. Also, exploring how subtle perturbations would result in significantly different outcomes can provide counterfactual insights into the clinical interpretation of model prediction. In this work, we introduce SurvAttack, a novel black-box adversarial attack framework leveraging subtle clinically compatible, and semantically consistent perturbations on longitudinal EHRs to degrade survival models' predictive performance. We specifically develop a greedy algorithm to manipulate medical codes with various adversarial actions throughout a patient's medical history. Then, these adversarial actions are prioritized using a composite scoring strategy based on multi-aspect perturbation quality, including saliency, perturbation stealthiness, and clinical meaningfulness. The proposed adversarial EHR perturbation algorithm is then used in an efficient SA-specific strategy to attack a survival model when estimating the temporal ranking of survival urgency for patients. To demonstrate the significance of our work, we conduct extensive experiments, including baseline comparisons, explainability analysis, and case studies. The experimental results affirm our research's effectiveness in illustrating the vulnerabilities of patient survival models, model interpretation, and ultimately contributing to healthcare quality.





\end{abstract}


\maketitle

\newcommand{\attack}{SurvAttack}

\input{tex/introduction}

\input{tex/methodology}

\input{tex/experiments}

\input{tex/related_work}

\input{tex/Conclusion}

\bibliographystyle{ACM-Reference-Format}
\balance
\bibliography{references}

\end{document}

%% file: tex/introduction.tex
\section{Introduction}



Mining health records uncovers valuable patterns and insights to enhance healthcare decision-making \cite{choi2016retain, hadizadeh2024contrastive, 10.1145/3709143, moghaddam2024meta}. Mining health records uncovers valuable patterns and insights to enhance healthcare decision-making \cite{choi2016retain, hadizadeh2024contrastive, 10.1145/3709143, moghaddam2024meta}. Survival analysis \cite{cox1972regression}, as a significant tool for time-to-event modeling, has been widely used in mining electronic health records (EHR) to analyze the time until the occurrence of an event of interest \cite{liu2018early, ohno2001modeling, katzman2018deepsurv, wang2019machine}. Despite extensive research on constructing more effective survival models \cite{wang2019machine}, there is a growing demand for model robustness and failure analysis when facing threatening situations given the high-stakes nature of health applications. Therefore, testing through a series of practical and effective ``adversarial attacks'' can be a necessary step for certifying the safety and robustness of a medical decision-support model before its deployment in the real world. One of the prevalent adversarial attacks is applying a curated imperceptible perturbation \cite{goldblum2022dataset} to input data, to ``trick'' a well-trained predictive model into overturning its decisions. In survival analysis, which estimates the patient's survival urgency and its underlying risk factors, the consequences of such attacks can be particularly severe as the patients in critical condition may lose their priority for ICU admission to the non-urgent patients. In such threatening situations, survival models are usually the ``black-box'' where adversaries lack knowledge of the model's architecture and parameters and cannot make any alterations to the model. Thus, the attack actions would solely fall on the input data. 

Studying adversarial attacks on survival models in this ``black-box'' setting can be beneficial in three ways.
First, this is one of the most practical means to initialize an attack which requires almost the lowest level of model access - inference. Therefore, the framework to study can account for a major portion of realistic threats.
Second, studying how these models get ``fooled'' by making minimal changes to the input is an effective approach to illustrate the inherent vulnerabilities of the model, prompting preemptive measures to avoid such high-risk failure in deployment.
Third, such attacks can offer explainability insights into the turning points and sensitive decision boundaries in the survival model’s decision-making process since they can pinpoint the critical positions along the EHR history. 
Although studies on adversarial attacks initially emerged in image recognition \cite{goodfellow2014explaining, papernot2016limitations, carlini2017towards, moosavi2016deepfool}, and later extended to more tasks such as natural language processing \cite{ebrahimi2017hotflip, zang2019word, ren2019generating}, the study of such attacks on healthcare modeling \cite{sun2018identify, an2019longitudinal, ye2022medattacker}, specifically for the task of survival models within the context of longitudinal health records, still remains highly unexplored. To bridge this research gap, there are two pressing challenges.



The first challenge is posed by the uniqueness of longitudinal EHR data. Generally, EHR data intrinsically exhibit a high-dimensional, discrete, and sparse feature space, where each patient instance consists of a temporal sequence of medical visits and each visit is associated with an unordered set of medical codes. Hence, unlike attacking image data where small amounts of numerical noise can be added to pixels, attacking EHR data takes an entire code as the smallest units of perturbation, such as removing or replacing an existing code or adding a new one. This different way of input perturbation introduces a higher difficulty of keeping the poisoned EHR sequence after manipulation clinically consistent with the original one. One naive solution is to count the number of codes being perturbed in a sequence, where more codes changed means lower clinical similarity before and after perturbation. However, given the diverse information each code carries and the varying time steps each code locates, choosing different codes to perturb would inevitably make an unequal impact on the ``health landscape'' of a patient. Therefore, how to keep the EHR perturbations aligned with the patient's health trajectory, and keep the perturbations conformed to clinical reality is a major challenge for designing adversarial attacks on EHR-based tasks. Otherwise, the required attack’s stealthiness is compromised.


The second challenge involves the distinct goal in survival analysis. Although survival analysis is generally seen as building a predictive model for ``time to event'', it is not merely a regression model because of the existence of censored data. The censored data refers to the patients whose event occurrences were not observed, for instance, due to the limited observation period or patient withdrawal during the study or hospitalization. Instead, we have a weaker label as ``length of survival'' since at least we know they survived (the event did not happen) till some point.
Therefore, on top of the regression task for observed patients (uncensored), survival models perform a ranking task to prioritize all patients (uncensored and censored) according to the predicted survival risk or duration. Due to serving a different goal from traditional supervised classification or regression tasks, survival models, a semi-supervised task, are evaluated by different metrics, such as the ranking-based c-index measuring how well the model can correctly rank a pair of patients by their relative survival risks. Therefore, how to effectively attack the input data not only to downgrade the predictive accuracy of ``time to event'', but also to disrupt the ranking quality of ``length of survival'' makes another major challenge in this study.

To this end, we propose SurvAttack, a novel black-box adversarial attack framework tailored for survival modeling in longitudinal health records, featuring the following contributions:

\begin{itemize}[leftmargin=*]

\item First, we design a greedy algorithm evaluating three adversarial actions over each medical code in an EHR sequence, based on a composite code
scoring (CCS) performing as a function of two elements. (1) the saliency index, reflecting the action's effect on the survival model's output; (2) the similarity index, indicating to what degree the post-perturbed EHR input holds similar semantic information as before clinically.

\item Second, we introduce an ontology-informed Synonym Code Selection (SCS) strategy to choose appropriate perturbation actions by ensuring their conformance to the clinical reality. SCS leverages domain knowledge using medical ontologies and statistical code co-occurrence probabilities. Also, the perturbations' stealthiness and medical consistency will be validated using a Semantic Similarity Function (SSF) which is enforced on the survival embedding space through an ontology-aware deep encoder. 

\item Third, the greedy EHR perturbation framework is finally used in a Dynamic SA-specific (DSA) strategy to efficiently attack a ``black-box'' survival model over the entire patient set. DSA dynamically manipulates the survival output for patients, aiming to increase or decrease the prediction of survival length, to maximally impair the model's ranking performance for patients' survival modeling.

\item Finally, we conduct an extensive experimental study on a large-scale real-world EHR dataset, by attacking multiple state-of-the-art survival models, which are comprehensively evaluated through baseline comparison, case studies, and attack pattern analysis for model interpretability. Both quantitative results and explanatory case studies demonstrate the utility of this work.
\end{itemize}

%% file: tex/methodology.tex
\section{Preliminary of Survival Analysis}
Survival models aim to estimate the survival function $S(t|V) = P_{V}(\tau>t)$ that represents the probability that an individual will survive beyond a specific time point $t$ given the subject's covariate $V$. In survival analysis, the mean lifetime denoted as $\mu$, representing the average time until an event occurs, is computed by integrating the survival function over time:
\begin{equation}
    \hat{T} = \int_{0}^{\infty} S(t|V) \, dt
\end{equation}
where $\hat{T}$ denotes the predicted survival time duration, and the output under attack in the victim survival models.

The objective of attacks on survival models is to undermine the survival model's ranking capability and compromise survival time prediction. Specifically, the following two survival analysis metrics are monitored to assess attack effectiveness in achieving its goals.

First, the Harrell’s Concordance Index (C-index) \cite{harrell1982evaluating}, which is used to assess concordance or agreement between the predicted survival times ranking and that of the actual survival times. A higher c-index indicates better concordance, meaning that the model's predicted rankings align more closely with the actual survival outcomes. 
Consider a pair of subjects $(i,j)$ with the actual survival times of $T_{i}$ and $T_{j}$, predicted survival times of  $\hat{T}_{i}$ and $\hat{T}_{j}$, and censoring labels of $k_{i}$ and $k_{j}$ (censored $k=0$, observed $k=1$). To compute the c-index, we consider three types of patient pairs: 
\begin{enumerate}[leftmargin=*,noitemsep,topsep=0pt]
\item Patients in the pair are both observed subjects ($k_{i} = k_{j} = 1$). We say $(i, j)$ is a concordant pair if $T_{j} > T_{i}$ and $\hat{T}_{j} > \hat{T}_{i}$ (correct ranking prediction), a discordant pair if $T_{j} > T_{i}$ and $\hat{T}_{j} < \hat{T}_{i}$ (incorrect ranking prediction), and a tied pair if $\hat{T}_{j} = \hat{T}_{i}$.

\item Patients in the pair are both censored subjects ($k_{i} = k_{j} = 0$). They are excluded from the computation because it's impossible to determine who experienced the event first (unrankable).

\item Patient $i$ is a censored subject and patient $j$ is an observed subject ($k_{i} = 0, k_{j} = 1$). If $T_{j} > T_{i}$, the pair is excluded because, again, it's unclear who experienced the event first. If $T_{j} < T_{i}$, $(i, j)$ is a concordant pair if $\hat{T}_{j} < \hat{T}_{i}$, a discordant pair if $\hat{T}_{j} > \hat{T}_{i}$, and a tied pair if $\hat{T}_{j} = \hat{T}_{i}$. 
\end{enumerate}

Given the defined pairs, Harrell’s C-index is defined as:
\begin{equation}
    C=\frac{\sum_{i, j} \left[\mathrm{I}\left(\hat{T}_i<\hat{T}_j\right) + 0.5\cdot\mathrm{I}\left(\hat{T}_i=\hat{T}_j\right) \right] \cdot \mathrm{I}\left(T_{i}<T_{j}\right) \cdot {k}_j}{\sum_{i, j} \mathrm{I}\left(T_{i}<T_{j}\right) \cdot {k}_i}
\end{equation}
which is the number of concordant pairs plus half the number of tied pairs, divided by the number of permissible (rankable) pairs. The operator $I(condition)$ is an indicator function that returns $1$ for true conditions and $0$ otherwise.

The second metric, Mean Absolute Error (MAE), measures the accuracy of time predictions for observed data:
\begin{equation}
    \text{MAE} = \frac{1}{N_\text{ob}} \sum_{i=1}^{N_\text{ob}}\|T_{i} - \hat{T}_{i}\|
\end{equation}
where $N_\text{ob}$ is the number of observed subjects.

\section{Proposed Method}


\begin{figure}[t]
    \begin{center}
    \centering
    \includegraphics[width=1.05\linewidth]{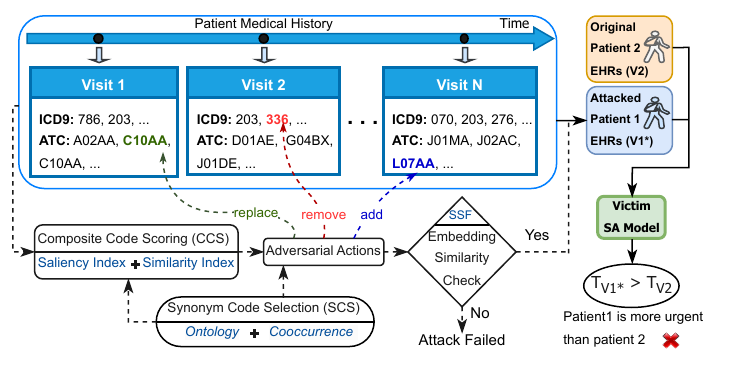}
    \captionsetup{skip=1pt}
    \caption{SurvAttack framework: Utilizing a greedy code manipulation algorithm, it assesses three potential adversarial actions (remove, replace, add) for each code using the CCS strategy, which integrates saliency and similarity elements, to disrupt the survival model's performance. The conformance of EHR perturbations to the clinical reality is maintained through the SCS strategy and the SSF function. As shown, manipulating the survival model’s ranking estimation 
 prioritized a non-urgent patient over a more critical one.}
    \label{fig:SurvAttack}
    \vspace{-0.3cm}
    \end{center}
\end{figure}

In this section, we first introduce the notations used in the paper. Subsequently, we present the details of our proposed method.

\subsection{Notation and Problem Definition}
Electronic Health Records (EHRs) contain comprehensive information about patients' medical history. Usually, for each patient with covariate $V$ representing the patient's medical history, multiple hospital visits $v_{n}$ have been recorded at each time point $n$ ($1 \leq n \leq N$), where $V =\{v_{n}\}_{n=1}^{N}$. In each visit $v_{n} =\{c_{i}\}_{i=1}^{s_{n}}$, there are $s_{n}$ number of standardized medical codes $c_{i}$ representing diagnoses, prescriptions, procedures, etc., which are documented within hospitals' databases. 
Three sets of information are needed from EHRs to enable survival analysis: 1) patient covariate $V$, 2) time of the event $T$, and 3) a label $k$ indicating the event observation status (censored $k=0$/observed $k=1$). Consider a survival model $F:\mathcal{V} \to \mathcal{T}$, which maps from input covariate space $\mathcal{V}$, containing all the possible EHR information of patients in the form of data $V$, to output survival time space $\mathcal{T}$ encompassing all the survival time $T \in \mathbb{R}$. We can feed a patient's covariate $V$ to $F$ to derive the estimated survival time $F(V) = \hat{T}$. Given two subjects of $V_{i}$ and $V_{j}$ with true labels of $(T_{i}, k_{i})$ and $(T_{j}, k_{j})$, the survival model $F$ strives to predict $\hat{T_{i}}$ and $\hat{T_{j}}$ to first be concordant with the ranking of actual survival times, meaning if $\mathrm{I}(T_{i}<T_{j}) \cdot k_{i} = 1$ then $\hat{T}_{i}<\hat{T}_{j}$, and second, if $k_{i}=1$, then $\hat{T_{i}} \approx T_{i}$, indicating accurate time prediction.

We attack the victim survival model $F$ by adding an indiscernible adversarial perturbation $\Delta V$ to one of the subjects (e.g. $i$) to create an adversarial subject $V^{*}_{i} = V_{i} + \Delta V_{i}$, tricking the victim model in two ways: first, it outputs a wrong ranking result $\hat{T_{i}^{*}}\nprec \hat{T_{j}}$ given $\hat{{T}_{i}}\prec\hat{T}_{j}$ (ranking flipped), and second, it enlarges the predictive error $|T_{i} - \hat{T_{i}^{*}}| > |T_{i} - \hat{T_{i}}|$ (accuracy deterioration). On the input side, the adversarial perturbation also requires to be subtle in a way that $V$ and $V^{*}$ are almost indistinguishable. We formulate this constraint by keeping the similarity of the $V$ and $V^{*}$ upper than a threshold: $SSF(V, V^{*})\geq\theta$, where $SSF$ is a domain-specific semantic similarity function presented in Section \ref{sec:SimFun}.
The threshold $\theta \in \mathbb{R}$ upholds the utility and semantic-preserving property of the perturbation. In other words, we adversarially manipulate patient $i$'s EHR input to obtain the corresponding manipulated survival time $\hat{T_{i}^{*}}$, flipping the correct ranking between $i$ and $j$. This manipulation deteriorates the survival model's performance in terms of concordance index and time prediction accuracy.



\subsection{Overview of the Proposed \attack}

In this section, we present an overview of our proposed black-box adversarial attack framework, named SurvAttack, targeting on EHR survival models. As shown in Figure \ref{fig:SurvAttack}, SurvAttack employs a \textbf{greedy search} that begins by first considering three adversarial action candidates (removal, replacement, and addition) for medical codes within each hospital visit of a patient's records. In the cases of replacing or adding, to maintain clinical consistency and conformity, we have devised a meticulous \textbf{ontology-informed synonym code selection (SCS) strategy} (section \ref{sec:SCS}) that integrates the medical domain knowledge and statistical information. This approach aims to ensure seamless alignment of any added code with the clinical semantic context in the patient's medical record. Also, we designed a \textbf{semantic similarity function (SSF)} (section \ref{sec:SimFun}) leveraging a SOTA ontology-aware deep encoder. SSF is utilized in both the adversarial action scoring and the ultimate attack stealthiness check. Using a \textbf{composite code scoring (CCS)} mechanism (section \ref{sec:CCS}), SurvAttack evaluates each of the adversarial candidates. These actions are subsequently sorted in descending order based on their associated composite scores. The attack (section \ref{sec:Attack}) proceeds in this order, executing high-scoring adversarial candidates first until either the adversarial attack succeeds or one of the attack constraints is breached (attack fails). Finally, we propose a \textbf{dynamic SA-specific attack (DSA)} strategy in section \ref{sec:attack_strategy}, which is specifically designed for survival analysis, dynamically using the defined greedy adversarial perturbation to disrupt the ranking ability of the SA model when performing on a set of patients.

\begin{figure}[t]
    \begin{center}
    \centering
    \includegraphics[width=1.0\linewidth]{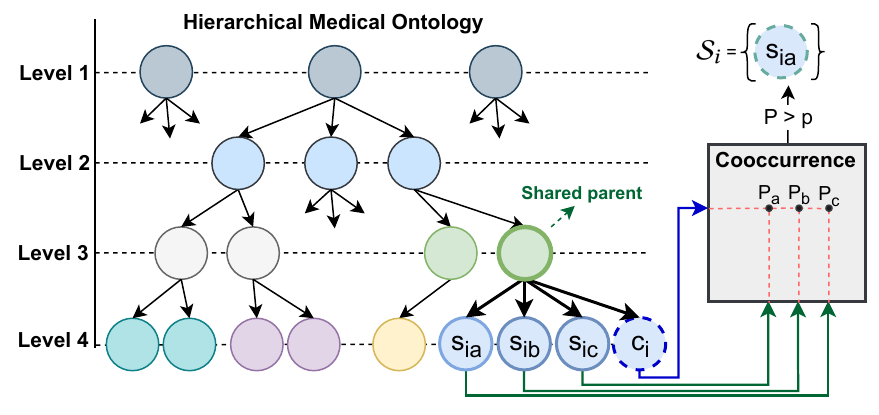}
    \captionsetup{skip=1pt}
    \caption{Ontology-informed Synonym Code Selection (SCS). SCS integrates the medical ontology and co-occurrence information to identify a similar set of codes, which are conformed to the clinical reality of the patient's EHR history.}
    \label{fig:SCS}
    \vspace{-0.3cm}
    \end{center}
\end{figure}

\subsection{Ontology-informed Synonym Code Selection (SCS)}\label{sec:SCS}

The Synonym Code Selection (SCS) strategy, as illustrated in Figure \ref{fig:SCS}, incorporates both medical ontological data as clinical domain knowledge and co-occurrence probabilities as overall statistics in datasets. This meticulous integration identifies a synonym set of codes for a targeted medical code for attack, thereby maintaining both clinical conformity and statistical consistency.

A medical ontology, depicted in Figure \ref{fig:SCS}, is a hierarchical structure of medical concepts, like a directed acyclic graph (DAG), that categorizes medical codes based on their similarities in clinical meaning. For example, In the International Classification of Diseases (ICD) coding system, codes within the same group represent related diseases, symptoms, or procedures. Similarly, in the Anatomical Therapeutic Chemical (ATC) classification system, the drug codes within the same categorization class share pharmacological or therapeutic properties. The ontology starts with broad general categories at the top level that encompass a wide range of related concepts and becomes increasingly specific as the classification descends. To choose an appropriate synonym code, we begin by identifying the targeted code's siblings. These are other codes within the same group under the same parent code, usually related and sharing semantic similarity. However, sometimes, two sibling codes, although sharing many similarities, may not occur simultaneously in reality. For instance, while the diagnoses ``Type I Diabetes Mellitus (ICD-9 code: 250.01)'' and ``Type II Diabetes Mellitus (ICD-9 code: 250.00)'' both stem from the same ancestor, ``Diabetes mellitus without complication (ICD-9 code: 250.0)'', they almost never co-occur in the EHR data as they have different underlying causes. It's more common for individuals to have either Type I or Type II diabetes. Another example is ``Fracture of neck of femur, closed (ICD-9 code: 820.8)'' and ``Fracture of neck of femur, open (ICD-9 code: 820.9)''. A fracture of the femur neck cannot be both open and closed simultaneously. These codes describe mutually exclusive conditions where a closed fracture means the skin is not broken, and an open fracture means there is a break in the skin. As a result, using such codes together or substituting one for another may not be consistent with the clinical reality of the patients and could compromise the stealthiness of the attack. Therefore, we add a second component which is code co-occurrence information to mitigate the aforementioned problem. A high co-occurrence probability between two medical codes indicates a strong statistical correlation and consistency. 
We utilized a co-occurrence conditional probability calculated as follows for the target code $c_{i}$ and its sibling $s_{ij}$:
\begin{equation}
    P[s_{ij},c_{i}] = Pr(s_{ij}|c_{i})
\end{equation}
  
We calculate the co-occurrence for all siblings of a targeted code and then selectively retain sibling codes with probabilities above a specified threshold ($P[s_{ij},c_{i}] > p$) and discard those below it. This process yields a set of similar codes 
$\mathcal{S}_{i}$, called synonym set, which represents the codes most similar to the original code $c_i$ considering medical knowledge and statistical information, ensuring the clinical meaningfulness and medical consistency of perturbations. 

\begin{figure}[t]
    \begin{center}
    \centering
    \includegraphics[width=1.0\linewidth]{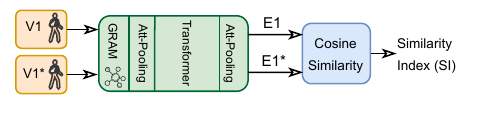}
    \captionsetup{skip=1pt}
    \caption{Semantic Similarity Function (SSF). SSF employs an ontological transformer-based encoder for embeddings, followed by cosine similarity to calculate the similarity index.}
    \label{fig:SimFunc}
    \vspace{-0.5cm}
    \end{center}
\end{figure}

\subsection{Semantic Similarity Function (SSF)}\label{sec:SimFun}
In many prior studies on adversarial attacks involving discrete EHR data \cite{ye2022medattacker, sun2018identify}, the enforcement of similarity between the attacked subject and the original version relied on the count of perturbations. However, this approach is inherently simplistic as it fails to consider the distinct impact of each action (e.g., removal, replacement, addition, etc.) on the semantic information within the patients’ health landscape. Moreover, even perturbations of the same action do not have the same semantic effect as each code carries a different amount of information given the nature of the event. To accurately evaluate the preservation of semantic information, we utilize a deep learning-based semantic similarity function (SSF), depicted in Figure \ref{fig:SimFunc}. Specifically, we employ an ontology-aware transformer-based encoder capable of extracting essential information and representing it within a distilled embedding vector $E = Enc(V) \in \mathbb{R}^{d}$. This encoder consists of three main blocks: 1) GRAM \cite{choi2017gram}, which utilizes the attention mechanism on ontology hierarchy, incorporating medical domain knowledge, 2) attention-pooling, compressing the data flow using attention mechanism, and 3) transformer \cite{vaswani2017attention}, harnessing the self-attention across multiple visits. This encoder was firstly pre-trained through a survival analysis target in work \cite{nayebi2023contrastive}. The survival predictor layer was then discarded after training. By feeding the encoder with the covariate \( V \) and its perturbed version \( V^{*} \), we infer the embedding vectors \( E \) and \( E^{*} \), then we use cosine similarity as the final step to derive the similarity index. The similarity function $SSF: \mathbb{R}^{n} \times \mathbb{R}^{n} \rightarrow \mathbb{R}$ is defined as:
\begin{equation}
    \begin{aligned}
    {SSF}(V, V^{*}) = Cos(Enc(V), Enc(V^{*})) =  \frac{{E \cdot E^{*}}}{{\|E\| \|E^{*}\|}}
    \end{aligned}
\end{equation}

The $SSF$ function will be utilized in the following Composite Code Scoring (CCS) section for code suggestion and in the Greedy EHR Adversarial Attack section for checking attack stealthiness.

\subsection{Composite Code Scoring (CCS)}\label{sec:CCS}
The adversarial attack begins with iterating over the codes inside the history of a patient and scoring them regarding possible adversarial actions. We consider only removal and addition during the scoring phase and bring the replacement action on top of the removal during the attack phase\footnote{Experimentally, we realized that this strategy yields better results compared to treating replace separately during the scoring phase}. We compute a saliency index and a similarity index for each adversarial action. The saliency index $\Delta F$ is the amount of change in the output of the victim survival model $F$ when a certain adversarial action is executed. Also, the similarity index $SI$, computed by the semantic similarity function (SSF), quantifies the semantic similarity level between the original $V$ and its perturbed version. In a visit $v_{n}$ of a patient's EHR $V$, taking the code $c_{i}$ as target, for removal, we have: 
\begin{align}
\Delta F_{i}^{r} &= F(V-c_{i}) - F(V)\\
SI_{i}^{r} &= SSF(V, V-c_{i})
\end{align}
where $\Delta F_{i}^{r}$ is the saliancy index regarding removal of code $c_{i}$, $V-{c_{i}}$ is the subject covariate when $c_{i}$ is removed, and $SI_{i}^{r}$ is the similarity index between $V_{i}$ and $V-{c_{i}}$. In terms of adding, first, we use the SCS strategy to find a set of synonym codes $\mathcal{S}_{i}$ for $c_{i}$. For each of the synonym codes $s_{ij}$ in $\mathcal{S}_{i}$ we have: 
\begin{align}
   \Delta F_{ij}^{a} &= F(V \cup s_{ij}) - F(V)\\
   SI_{ij}^{a} &= SSF(V, V \cup s_{ij})
\end{align}
where $\Delta F_{ij}^{a}$ is the saliancy index regarding adding of code $s_{ij}$ as a sibling code to $c_{i}$ from ontology information, $V \cup s_{ij}$ is the manipulated covariate when $s_{ij}$ is added, and $SI_{ij}^{a}$ is the similarity index between $V$ and $V \cup s_{ij}$. 

\begin{algorithm}[t]
    \caption{Composite Code Scoring (CCS)}\label{alg1}
    \small
    \begin{algorithmic}[1]

        \State
        \textbf{Input:} Patient covariate $V$, Victim survival model $F$

        \State
        \textbf{Output:} List of ordered EHR codes $\mathcal{C}_{ord}$

        \State Initialize $\mathcal{C} \gets \{\}$ 

        \For{$v_{n}$ in $V$}        
            \For{$c_{i}$ in $v_{n}$} 
                \State label $l_{i} \gets \text{``remove''}$
                \State Compute saliency $\Delta F_{i}^{r}$, and similarity $SI_{i}^{r}$ indexes 
                \State Compute composite score $h^{r}_{i} = \mathcal{H}(\Delta F_{i}^{r}, SI_{i}^{r})$
    
                \State Add element to $\mathcal{C}$: $(c_{i}, v_{n}, l_{i}, h^{r}_{i})$ 
    
                \State Compute $\mathcal{S}_{i}$ using \textbf{Synonym Code
Selection (SCS)} 
    
                \For{$s_{ij}$ in $\mathcal{S}_{i}$}
                    \If{$s_{ij}$ not in $V$}
                        \State label $l_{ij} \gets \text{``add''}$
                        \State Compute saliency $\Delta F_{ij}^{a}$, and similarity $SI_{ij}^{a}$ indexes  
                        \State Compute composite score $h^{a}_{ij} = \mathcal{H}(\Delta F_{ij}^{a}, SI_{ij}^{a})$
                        \State Add element to $\mathcal{C}$: $(s_{ij}, v_{n}, l_{i}, h^{a}_{ij})$ 
                    \EndIf
                \EndFor
            \EndFor
        \EndFor

        \State $\mathcal{C}_{\text{ord}} \gets Sort(\mathcal{C})$ based on composite score $h$

    \end{algorithmic}
\end{algorithm}

Finally, we combine saliency index $\Delta F$ and similarity index $SI$ into a composite score $h$, serving as the overall sorting criterion, and represented as the function $H$ of $\Delta F$ and $SI$:
\begin{equation}
\label{eq1}
    h = \mathcal{H}(\Delta F, SI)= \Delta F \cdot \phi(SI)
\end{equation}
where 
    $\phi(z) = e^{\lambda z}$ 
and $\lambda$ is a hyperparameter. The function $\phi$ exponentially amplifies scores for higher similarities and rapidly diminishes them for lower scores. The inclusion of similarity in our sorting strategy aims to prioritize actions that maintain semantic similarity early in the attack sequence while penalizing those that compromise it, placing them later. This approach enhances attack performance by reducing the risk of violating the similarity constraint (leading to attack failure) while simultaneously altering the output more substantially.

Considering all potential adversarial actions for existing codes, we assign each one a composite score $h$, and one label $l$ indicating the type of action. Finally, we sort them in descending order regarding their composite scores. This arrangement is used for a targeted attack aimed at increasing the predicted survival time for the patient. Conversely, to deceive the victim model into predicting a lower survival time, we can invert the sign of composite scores by multiplying them by $-1$ before sorting. Algorithm \ref{alg1} illustrates the composite code scoring (CCS) strategy.

\begin{algorithm}[t]
    \caption{Greedy EHR Adversarial Attack}\label{alg2}
    \small
    \begin{algorithmic}[1]
        \State \textbf{Input:} Patient covariate $V$, Victim survival model $F$, Target time $t$
        \State \textbf{Output:} Adversarial patient covariate $V^{*}$

        \State Compute original predicted survival time $F(V)=\hat{T}$


        \State $\text{direction} \gets \text{``increase'' if } \hat{T} < t \text{ else ``decrease''}$

        
        \State $\mathcal{C}_{\text{ord}} \gets$ \textbf{Alg. \ref{alg1} Composite Code Scoring (CCS)}

        \State Initialize $V^{*} \gets V$

        \For{$(c_{i}/s_{ij}, v_{n}, l_{i}, h_{i}^{r/a})$ in $\mathcal{C}_{\text{ord}}$}
        
            \State Compute current predicted survival time $F(V^{*})=\hat{T}_{cur}$ 
\Comment{ \hspace{1em}\(\blacktriangleright\)---------------\textit{Adversarial action: adding}---------------}            
            \If{$l== \text{add}$}

                \State Add $s_{ij}$: $V^{*} \gets (V^{*}\cup s_{ij})$ 

                \If{$\Delta F = F(V^{*}) - \hat{T}_{cur}$ towards direction}
                    \State keep the added code
                        \If{$\textbf{SSF}(V, V^{*}) < \theta$}
                            \State reverse the adding
                            \State Break
                        \EndIf
                \EndIf
\Comment{ \hspace{1em}\(\blacktriangleright\)---------------\textit{Adversarial action: remove}---------------}    
            \Else 
                \State Remove $c_{i}$: $V^{*} \gets (V^{*}-c_{i})$
                \If{$\Delta F = F(V^{*}) - \hat{T}_{cur}$ towards direction}
                    \State Keep the removal action
                        \If{$\textbf{SSF}(V, V^{*}) < \theta$}
                            \State reverse the removal
                        \EndIf
                \EndIf
\Comment{\hspace{2.2em} \(\blacktriangleright\)---------------\textit{Adversarial action: replace}---------------}
            \State Fetch $\mathcal{S}_{i}$ of $c_{i}$ calculated in CCS
            \State Remove synonym codes from ${S}_{i}$ that already exists in $V$

            \State $s_{ij}^{*} = \arg \max _{s_{ij} \in \mathcal{S}_{i}}\left\{\mathcal{H}(V^{*}, (V^{*} - c_{i}) \cup s_{ij})) \right\}$
            \State Replace $s_{ij}^{*}$ with $c_{i}$: $V^{*} \gets ((V^{*}-c_{i})\cup s_{ij})$ 

            
            \If{$\Delta F = F(V^{*}) - \hat{T}_{cur}$ towards direction}
                \State keep the replacement 
                    \If{$\textbf{SSF}(V, V^{*}) < \theta$}
                        \State Reverse the replacement 
                        \State Break
                    \EndIf
            \EndIf
            
            \EndIf

        \EndFor

    \end{algorithmic}
\end{algorithm}
\vspace{-0.2cm}

\subsection{Greedy EHR Adversarial Attack}\label{sec:Attack}

We start the adversarial attack by initializing the adversarial subject $V^{*}$ with $V$. Using the CCS strategy to identify and rank all potential adversarial actions, we begin with executing the highest-scored actions first. If the action's label $l$ is "add", we consider adding the code to $V^{*}$, and calculate the model's output change $\Delta F$ and similarity index $SI$. While not violating the similarity threshold, if the output of the model changes to the desired direction ($\Delta F >0$ for increase, and $\Delta F <0$ for decrease), we keep the added code. In the case of violating the similarity threshold, we drop the added code and stop the process. 

If the label $l$ is "remove", we consider removing the code and repeating the same steps of the model's output change and similarity check. Then, we consider replacing the code with a similar code from the synonym set $\mathcal{S}_{i}$ which was already created in the scoring phase. In this stage, we calculate the score of each replacement candidate using the composite score $\mathcal{H}$ and choose the code from synonym set that has the highest score.
\begin{equation}
   s_{ij}^{*} = \operatorname*{argmax} _{s_{ij} \in \mathcal{S}_{i}}\left\{ \mathcal{H}(V^{*}, (V^{*} - c_{i}) \cup s_{ij})) \right\} 
\end{equation}
We execute the replacement step and subsequently proceed with the output and similarity checks, as previously described. We compare the removal and replacement and proceed with the action that yields the greatest model's output change while maintaining similarity. We note that if the similarity criterion is not met during the removal phase, we do not halt the process and instead continue to the replacement step to see if the similarity comes back above the threshold. The attack continues until the model's output is successfully altered to the desired value (flipping the predicted survival ranking) or fails, such as when breaching the similarity threshold. Algorithm \ref{alg2} shows the attack details.

\begin{algorithm}[t]
    \caption{Dynamic SA-specific Attack (DSA) Strategy}\label{alg3}
    \small
    \begin{algorithmic}[1]
        \State \textbf{Input:} Patient set $\mathcal{P} = {V_1, ..., V_{|\mathcal{P}|}}$
        \State \textbf{Output:} Adversarial patient set $\mathcal{P}^{*} = {V^*_1, ..., V^*_{|\mathcal{P}|}}$

        \State $\mathcal{P}^{\text{ob}},\mathcal{P}^{\text{cen}} \gets$ Dividing observed and censored patients from $\mathcal{P}$
        \State $\mathcal{P}^{\text{ob}}_{\text{ord}} \gets Sort(\mathcal{P}^{\text{ob}})$ according to true survival time $T$

        \State Initialize $\mathcal{P}^{\text{cen}*} \gets \{\}$
        \For{$V^{\text{cen}}_{i}$ in $\mathcal{P}^{\text{cen}}$}
            \State $V^{\text{cen}*}_{i} \gets $ Applying \textbf{Alg. \ref{alg2}} with target time $t=0$ on $V^{\text{cen}}_{i}$ 
            \State Add element to $\mathcal{P}^{*}$: $V^{\text{cen}*}_{i}$
        \EndFor

        \State Initialize $\mathcal{P}^{\text{ob}*} \gets \{\}$, $t_{min} \gets$ max $\hat{T}^{\text{cen}}$ in $\mathcal{P}^{\text{cen}*}$ 

        \For{$V^{\text{ob}}_{i}$ in $\mathcal{P}^{\text{ob}}_{\text{ord}}$}
            \State $V^{\text{ob}*}_{i} \gets $ Applying \textbf{Alg. \ref{alg2}} with target time $t=t_{min}$ on $V^{\text{ob}}_{i}$
            \State Add element to $\mathcal{P}^{\text{ob}*}$: $V^{\text{ob}*}_{i}$
            \State $t_{min} = F(V^{\text{ob}*}_{i})$
        \EndFor
        \State $\mathcal{P}^{*} \gets Merge(\mathcal{P}^{\text{ob}*}, \mathcal{P}^{\text{cen}*})$

    \end{algorithmic}

\end{algorithm}
\vspace{-0.2cm}

\begin{table*}[ht]
\renewcommand*{\arraystretch}{1.1}
\setlength{\tabcolsep}{10pt}
\centering
\small
\caption{Baseline comparison for MAE and c-index in various cases. }
\vspace{-0.1cm}
    \label{table:surv-attack}
    \setlength{\tabcolsep}{4pt} 
    \begin{tabular}{c|c c c c c|c c c c c}
        \hline 
        \multirow{2}{*}{\textbf{Attack Method}} &
        \multicolumn{5}{c|}{\textbf{CoxCC}} & \multicolumn{5}{c}{\textbf{DeepSurv}} \\
        \cline{2-11}
        & $\textbf{c}_{1}$ & $\textbf{c}_{2}$ & $\textbf{c}_{t}$ & $\textbf{c}_{ob}$ & $\textbf{MAE}$ & $\textbf{c}_{1}$ & $\textbf{c}_{2}$ & $\textbf{c}_{t}$ & $\textbf{c}_{ob}$ & $\textbf{MAE}$ \\
        \hline
        No Attack & 0.7107 & 0.7107 & 0.7107 & 0.5485 & 1.83 & 0.6934 & 0.6934 & 0.6934 & 0.5504 & 2.01 \\
        \hline
        Random & 0.742 & 0.6064 & 0.6405 & 0.5369 & 2.00 & 0.7814 & 0.4481 & 0.5636 & 0.5051 & 2.79 \\   
        Surv-TextBugger & 0.3266 & 0.2722 & 0.1132 & 0.4997 & 2.98 & 0.2520 & 0.1946 & 0.08615 & 0.4857 & 3.89\\
        Surv-WS & 0.4986 & 0.5475 & 0.3292 & 0.5136 & 2.12 & 0.4028 & 0.4542 & 0.2215 & 0.5153 & 2.78\\
        Surv-PWWS & 0.3973  & 0.1808  & 0.098  & 0.5074 & 3.39  & 0.2827  & 0.1191  & 0.0738  & 0.4684 & 4.27 \\
        SurvAttack-NoSym  & 0.3722 & 0.1701  & 0.0918  & 0.4753 & 3.68 & 0.2745 & 0.1024  & 0.0721  & 0.4181 &  4.31 \\        
        \textbf{SurvAttack} & \textbf{0.3102} & \textbf{0.1489} & \textbf{0.0841} & \textbf{0.4552} & \textbf{3.93} & \textbf{0.2057} & \textbf{0.0889} & \textbf{0.0569} & \textbf{0.4126} & \textbf{4.48} \\        
        \hline
        
        \multirow{2}{*}{\textbf{Attack Method}} & \multicolumn{5}{c|}{\textbf{N-MTLR}} & \multicolumn{5}{c}{\textbf{DeepHit}} \\
        \cline{2-11}
        & $\textbf{c}_{1}$ & $\textbf{c}_{2}$ & $\textbf{c}_{t}$ & $\textbf{c}_{ob}$ & $\textbf{MAE}$ & $\textbf{c}_{1}$ & $\textbf{c}_{2}$ & $\textbf{c}_{t}$ & $\textbf{c}_{ob}$ & $\textbf{MAE}$ \\
        \hline
        No Attack & 0.5170 & 0.6242 & 0.6242 & 0.6242 & 3.49 & 0.5589 & 0.6527 & 0.6527 & 0.6527 & 2.06 \\  
        \hline
        Random & 0.4997 & 0.7003 & 0.4747 & 0.5613 & 4.69 & 0.5056 & 0.7917 & 0.3318 & 0.5169 & 2.48 \\        
        Surv-TextBugger & 0.4405 & 0.3201 & 0.1918 & 0.0111 & 5.44 & 0.4873 & 0.2001 & 0.1551 & 0.078 & 3.08 \\
        Surv-WS & 0.4844 & 0.5318 & 0.3136 & 0.3045 & 4.93 & 0.5012 & 0.2842 & 0.16 & 0.0915 & 2.65\\
        Surv-PWWS & 0.4880 & 0.5460 & 0.2783  & 0.3020  & 5.17  & 0.4509 & 0.2154  & 0.0794  & 0.0645  & 3.04 \\
        SurvAttack-NoSym & 0.4270 & 0.2841 & 0.2987  & 0.0993  & 5.53  & 0.5026 & 0.2042 &  0.0698 & 0.0668  & 3.32  \\
        \textbf{SurvAttack} & \textbf{0.3926} & \textbf{0.2733} & \textbf{0.0588} & \textbf{0.0595} & \textbf{5.72} & \textbf{0.4809} & \textbf{0.1417} & \textbf{0.0688} & \textbf{0.0644} & \textbf{3.91} \\
        \hline
    \end{tabular}
    \vspace{-0.2cm}
\end{table*}

\subsection{Dynamic SA-specific Attack (DSA) Strategy for Population Survival Disruption}\label{sec:attack_strategy}



Finally, Given SurvAttack's primary objective, disrupting the predicted survival ranking of patients inside a patient cohort in the hospital, we design a dynamic SA-specific (DSA) strategy to adversarially adjust patients' predicted survival time, aiming to convert all permissible pairs to be discordant (a patient pair with incorrect survival ranking). This ultimately results in a low c-index and high MAE. Unlike a naive approach of attacking all possible permissible pairs to flip their concordance, DSA achieves the goal with a single attack per patient, minimizing attack computations.

In the first step, we want to target all observed-censored concordant pairs $(k_{i}=1, k_{j}=0)$ by attacking censored subjects to reduce their predicted survival time, aiming to flip the correct ranking of these pairs. Before the attack, we have $\mathrm{I}(\hat{T}_i<\hat{T}_j) \cdot \mathrm{I}(T_{i}<T_{j})=1$, but after the attack we have $\hat{T}_i>\hat{T}_j$, so $\mathrm{I}(\hat{T}_i<\hat{T}_j) \cdot \mathrm{I} (T_{i}<T_{j})=0$, meaning that the pair becomes discordant.

Next, we focus on observed-observed pairs. We set a starting target point $t_{min}$, initialized with the highest post-attack predicted survival time among censored subjects. Then, sorting observed subjects based on their survival time in ascending order, we start attacking them one by one with respect to target time $t_{min}$. 
If the observed patient's original predicted survival time $\hat{T}_{i}$ is less than $t_{min}$ ($\hat{T}_{i}<t_{min}$), we perturb it so that its predicted survival time is above $t_{min}$ ($\hat{T}^{*}_{i}>t_{min}$) and set $t_{min}=\hat{T}^{*}_{i}$. If the observed patient's original predicted survival time $\hat{T}_{i}$ is larger than $t_{min}$ ($\hat{T}_{i}>t_{min}$), we adversarially perturb it till the predicted survival time becomes closest to the $t_{min}$ but above it ($\hat{T}_{i}>t_{min}$), and then set $t_{min}=\hat{T}^{*}_{i}$. Performing this strategy, we ensure that ideally all the observed subjects' predicted survival times are above those of censored subjects (keeping observed-censored pairs discordant), while the survival ranking between each observed-observed pair is incorrect. DSA is detailed in Algorithm \ref{alg3} and also, Figure  \ref{fig:strategy} shows the visualization of this strategy.

%% file: tex/experiments.tex
\begin{figure*}[t]
    \begin{center}
    \centering
    \hspace*{-0.6cm}
    \includegraphics[width=0.95\linewidth]{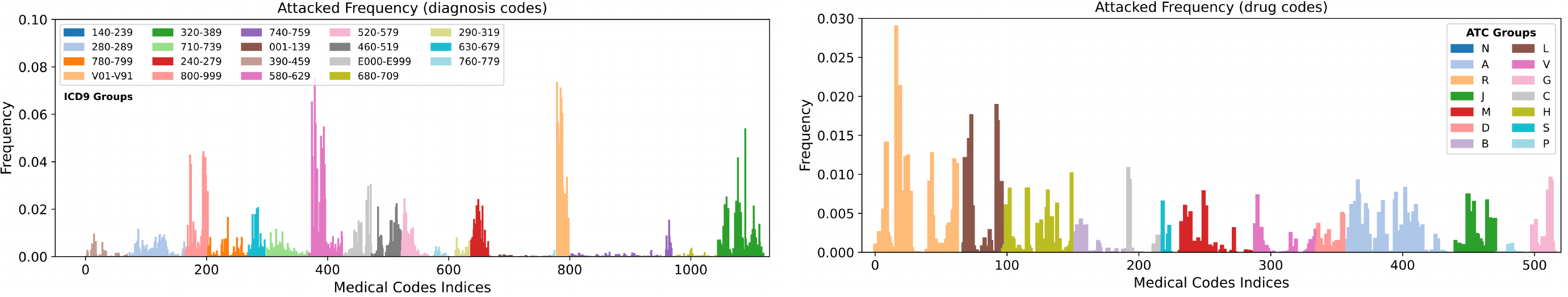}
    \captionsetup{skip=1pt}
    \caption{Frequency of adversarial attacks on Diagnosis Codes (left) and drug codes (right) executed by SurvAttack. Codes belonging to the same group in the specific ontology are distinguished by different colors.}
    \label{fig:fr-dx}
    \vspace{-0.3cm}
    \end{center}
\end{figure*}

\begin{figure*}[t]
    \begin{center}
    \centering
    \includegraphics[width=1.0\linewidth]{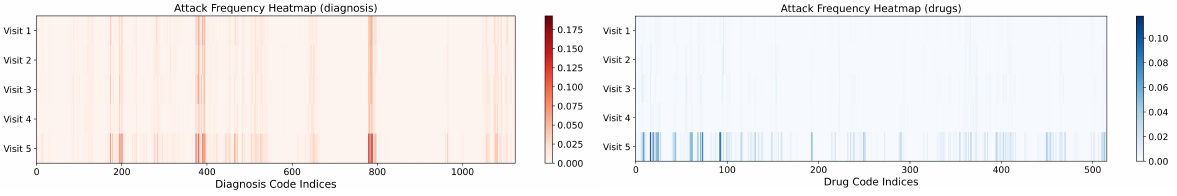}
    \captionsetup{skip=1pt}
    \caption{SurvAttack's perturbation patterns for diagnosis codes (left) and drug codes (right). These heatmaps illustrate the percentage of attacks on each code (x-axis) across visits (y-axis), with darker areas indicating a higher frequency of attacks.}
    \label{fig:attack-pos}
    \vspace{-0.3cm}
    \end{center}
\end{figure*}


\begin{table}[H]
\small
\centering
\captionsetup{skip=3pt}
\caption{Dataset Statistics}
\label{tab:Statistical}
\renewcommand{\arraystretch}{1.3}
\begin{tabular}{R{4.0cm} C{2.9cm}}
\toprule
\# of patients & 77809 \\
\# of censored patients & 65437 (84.1\%) \\
\# of observed patients & 12372 (15.8\%) \\
Patient average age & 59.62 \\
Sex distribution & (52\% M, 48\% F)\\
Average \# of codes per patient & 66.46 \\
\# of visits per patient & 5 \\
\hline
\end{tabular}
\vspace{-0.1cm}
\end{table}

\section{Experiments}
\subsection{Dataset Description}
We utilized a real-world Electronic Health Record (EHR) dataset  \cite{liu2022development} obtained from the Anonymous University Medical Center, encompassing the data of more than 77-thousand patients collected from early 2009 to late 2021, focusing on forecasting a critical and widely existing condition - Acute Kidney Injury (AKI)\footnote{This dataset is acquired from the hospital following stringent safety protocols and with proper authorization from relevant authorities.}. Compared to the public datasets, AKI offers a richer and longer history of visits for each patient (comparing to ICU stays in MIMIC), and a clearer labeling of onset dates, making it more suitable for survival analysis and attack study. More detailed statistics of the dataset are shown in Table \ref{tab:Statistical}.

\begin{figure}[t]
    \begin{center}
    \centering
    \includegraphics[width=0.85\linewidth, height=0.6\linewidth]{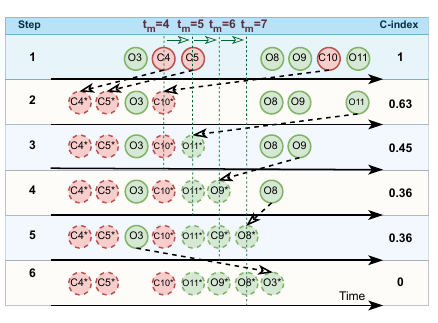}
    \captionsetup{skip=1pt}
    \caption{Dynamic SA-specific Attack (DSA) Strategy. Step 1: DSA attacks all the censored subjects to reduce their expected survival times, (disrupting concordance in observed-censored pairs) and then Initialize $t_{min}$ with the highest post-attack expected survival time of censored subjects. Step 2-6: Using $t_{min}$ as the target time, the algorithm proceeds by attacking the sorted observed subjects and updates $t_{min}$ accordingly (disrupting concordance in observed-observed pairs).}
    \label{fig:strategy}
    
    \end{center}
    \vspace{-0.3cm}
\end{figure}

\subsection{Experimental Setting}
For experimenting with SurvAttack, we set the co-occurrence ratio $p$ to 0.75 and similarity threshold $\epsilon$ to 0.90 and then attacked four popular survival models evaluated on a target set consisting of 3549 patient subjects (2984 censored and 565 observed). These survival models are as follows: (1) \textbf{CoxCC} \cite{JMLR:v20:18-424}: CoxCC uses neural networks to parameterize the relative risk function, allowing for the modeling of complex relationships between covariates and event times. (2) \textbf{DeepSurv} \cite{katzman2018deepsurv}: DeepSurv is a personalized treatment recommendation system based on a Cox proportional hazards deep neural network. (3) \textbf{N-MTLR} \cite{fotso2018deep}: The Neural Multi-Task Logistic Regression is founded on the Multi-Task Logistic Regression (MTLR) model, as introduced by Yu et al. \cite{yu2011learning}, with a deep learning architecture at its core. (4) \textbf{DeepHit} \cite{lee2018deephit}: DeepHit is a deep learning survival analysis method that employs multi-task learning to estimate survival time and event type probabilities, addressing competing risks. Trained on a set of 69599 patients, their evaluation results before the attack are demonstrated in Table \ref{table:surv-attack}.

Moreover, as for medical ontology used in synonym selection strategy (SCS), we employed the International Classification of Diseases (ICD-9) for diagnosis codes and the Anatomical Therapeutic Chemical (ATC) classification system for prescription codes. The anonymous GitHub implementation of SurvAttack\footnote{https://anonymous.4open.science/r/SurvAttack-4D0F/} is released.

Regarding computational complexity, if \( N \) is the total number of visits of a patient, \( S \) is the total number of codes in a visit, and \( C \) is the maximum number of acceptable synonyms for a code, the time complexity for attacking one patient is \( O(NSC) \). On average, our method requires approximately 18 seconds per attack (Hardware: NVIDIA RTX 6000 GPU, 256 GB of RAM, 64-core CPU), though this can vary based on factors such as computational resources, encoder complexity, survival model, and patient history length. 

\subsection{Results and Discussion}
As this is the first work designing an adversarial attack on health survival models, we choose some of the popular black-box baselines from other domains, mostly text classification, and adapt them for attacking survival models in EHRs. These adapted baselines are prefixed with "Surv". The five baselines are: (1) \textbf{Random}: A naive approach that randomly selects adversarial actions, without employing any survival analysis-specific strategy. (2) \textbf{Surv-PWWS}: Adapted from the probability weighted word saliency (PWWS) \cite{ren2019generating} which is a black-box attack algorithm. (3) \textbf{Surv-TextBugger}: Adapted from a black-box adversarial attack method referred to as TextBugger \cite{li2018textbugger}. (4) \textbf{Surv-WS}: Adapted from the black-box adversarial attack algorithm in \cite{samanta2017towards}. (5) \textbf{SurvAttack-NoSym}: Same as SurvAttack but excludes the similarity index in the composite score. Codes are sorted solely based on the saliency index. 

We compare the adversarial attack performance results of SurvAttack with the five introduced attack baselines, as shown in Table \ref{table:surv-attack}. The table shows the c-index and MAE before and after the attack regarding different cases. In a real scenario when deploying SA models in hospitals, the attacker knows which patients are more urgent maybe based on early medical discretion or prior accurate survival analysis. The attacker aims to disrupt the survival ranking of the SA model, prioritizing non-urgent patients over critical ones. This is similar to when all the subjects are observed. However, there might be cases when the survival urgency of some patients has not been determined yet (both censored and observed). 
Hence, we consider different cases in our evaluation to cover all potential scenarios, and also as an ablation study of different steps of the DSA strategy. We consider cases when we have both observed and censored patients but we attack only censored subjects ($c_{1}$), only observed subjects ($c_{2}$), and finally all the patients ($c_{t}$). Also, we consider the case when all the subjects are observed, and the entire set is targeted for attack ($c_{ob}$). SurvAttack outperforms other black-box adversarial attack baselines on four victim survival models regarding various evaluation cases, showcasing its superior performance. As expected, Random demonstrates the worst results. Comparing SurvAttack with SurvAttack-NoSym, we found that including similarity in the scoring phase improves attack performance by degrading both survival ranking and time prediction. Among the victim survival analysis models, when getting attacked by SurvAttack, DeepSurv experienced the greatest performance drop, with a 63.56\% reduction in c-index, while MTLR showed the least drop at 56.47\%. Considering $c_{1}$ and $c_{2}$ as the ablation study of DSA, we can see that mostly perturbing observed data plays a greater role in impairing the survival ranking performance.

\begin{table*}[ht]
\setlength{\tabcolsep}{3.5pt}
    \centering
    \footnotesize
    \caption{Top-$k$ most frequent attacked codes for each action averaged among attacks to the four survival models.}
    \vspace{-0.2cm}
    \label{tab:code_freq}
    \begin{tabular}{c|c c c c|c c c c|c c c c}
        \hline
        \multirow{2}*{$k$} & \multicolumn{4}{c|}{\textbf{Add}} & \multicolumn{4}{c|}{\textbf{Remove}} & \multicolumn{4}{c}{\textbf{Replace}} \\
        \cline{2-13}

         & Diagnosis (ICD9) & Fr(\%) & Drugs (ATC) & Fr(\%) & Diagnosis (ICD9) & Fr(\%) & Drugs (ATC) & Fr(\%) & Diagnosis (ICD9) & Fr(\%) & Drugs (ATC) & Fr(\%)\\
        \hline
        1 & \thead{410\\(heart attack)} & 12.4 & \thead{N02AJ (opioids+\\non-opi analg)} & 9 & \thead{780\\(general Sx.)} & 3.1 & \thead{B05XA\\(electrolyte)} & 3.8 & \thead{271 (carb.\\metabolism)} & 2.9 & \thead{B05XX \\(IV solution)} & 3.3 \\[-0.8ex]
        
        2 & \thead{411 (acute\\ischemic heart)} & 9.5 & \thead{B06AA\\(enzymes)}& 8.7 & \thead{401 (hypertension)} & 2.6 & \thead{B01AB (Heparin)} & 2.8 & \thead{782 (integumentary\\tissue Sx)} & 2.3 & \thead{N02AC\\(DiphenylprA)} & 2\\[-0.8ex]
        
        3 & \thead{275 (mineral\\metabolism)} & 6.8 & \thead{G02AD\\(prostaglandins)} & 6.4 & \thead{V58 (aftercare)} & 2.5 & \thead{D01AC (Imidazole,\\triazole)} & 2.5 & \thead{786 (respiratory\\system Sx.)} & 2.2 & \thead{N01BX (local\\anesthetics)} & 2\\[-0.8ex]
        
        4 & \thead{251 (pancreatic\\internal secretion)} & 6.4 & \thead{A11DA\\(vitamin B1)} & 4.7 & \thead{427 (heart\\failure)} & 2 & A06AA (emollients) & 2 & \thead{781 (nervous,\\musculoskeletal Sx.)} & 2 & \thead{A06AX\\(constipation rx)} & 1.8\\[-0.8ex]
        
        5 & \thead{422 (acute\\myocarditis)} & 5.5 & \thead{B02AB (protease\\inhibitors)} & 4.5 & \thead{787 (nausea)} & 2 & \thead{N01AH (opioid\\anesthetics)} & 2 & \thead{789 (abdomen,\\pelvis Sx.)} & 1.9 & \thead{B05XB\\(amino acids)} & 1.6\\
        \hline
    \end{tabular}
    \vspace{-0.2cm}
\end{table*}

\subsubsection{Adversarial Attack shed light on survival decision-making}
Table \ref{aa-visit} displays the attack percentage for each visit, averaged across attacks to all four victim survival models. Visit 5 stands out as the most targeted, underscoring its significant impact on the final output of these models, especially in terms of drug codes. Furthermore,  we present the top 5 most frequently targeted codes in Table \ref{tab:code_freq}, averaged among attacks to all four victim survival models. This table highlights the significance of each medical code in survival model decision-making. For example, the "410" ICD-9 code, as the most added code (existed in 12.4\% of attacks), represents acute myocardial infarction (heart attack). According to \cite{kanic2019acute, shacham2016acute} individuals experiencing a myocardial infarction (MI) face a 2.6-fold elevated risk of developing acute kidney injury (AKI). Figure \ref{fig:fr-dx} further elucidates the decision-making of survival models by depicting the attack percentage of all diagnosis and drug codes. Each ICD-9 group (on the left) and ATC group (on the right) is uniquely colored to provide a broader perspective on the significance of diagnoses and prescriptions in AKI time modeling. For instance, genitourinary system-related diseases (580-629) are among the primary targets of SurvAttack, reflecting their crucial role in survival decision-making, which makes sense since AKI is categorized as a genitourinary disorder. 
Also, the most-targeted drug codes are respiratory-related drug codes (R category in ATC), which shows that respiratory complications are common in patients with AKI \cite{faubel2016mechanisms, park2021acute, alge2021two}. Figure \ref{fig:attack-pos} displays perturbation patterns across time and EHR features (code). The x-axis represents code IDs, while the y-axis shows chronological visits. Each heatmap illustrates how SurvAttack perturbs sequential EHR data, with darker shades indicating more perturbations in terms of the percentage of being targeted. Evidently, the last visit is the primary focal point for attacks, particularly concerning drug codes. Therefore, it becomes the key region of interest in the time modeling of AKI within survival models.

\begin{table}[t]
\renewcommand*{\arraystretch}{1.1}
    \centering
    \small
    \caption{Case study. Adversarial attack to a patient with $T=6.86$, $\hat{T}=6.032$, $\hat{T}^{*}=4.318$, $SI=0.96\%$.}
    \vspace{-0.2cm}
    \begin{tabular}{c|c c c c c|c}
        \hline
        \textbf{Visit} & \textbf{1} & \textbf{2} & \textbf{3} & \textbf{4} & \textbf{5} & \textbf{Total} \\
        \hline
        \textbf{\# Codes} & 1 & 3 & 1 & 5 & 34 & 44 \\
        \hline
        \textbf{\# Perturbations} & 1 & 0 & 0 & 3 & 7 & 11 \\
        \hline
        \textbf{\# Rem/Rep/Add} & 0/1/0 & 0/0/0 & 0/0/0 & 1/1/1 & 1/2/4 & 2/4/5 \\
        \hline
    \end{tabular}
    \label{tab:case-study}
    \vspace{-0.4cm}
\end{table}

\begin{table}[H]
\renewcommand*{\arraystretch}{1.1}
    \centering
    \small
        \setlength{\tabcolsep}{4pt} %
    \caption{Attack percentage on different visits averaged among four victim survival models.}
    \vspace{-0.2cm}
    \label{aa-visit}
    \begin{tabular}{c|c c c|c c c|c c c}
        \hline
        \multirow{2}*{\textbf{visit}}&\multicolumn{3}{c|}{\textbf{Add (\%)}} & \multicolumn{3}{c|}{\textbf{Remove (\%)}} & \multicolumn{3}{c}{\textbf{Replace (\%)}} \\
        \cline{2-10}
         & \textbf{Diag.} & \textbf{Drug} & \textbf{All} & \textbf{Diag.} & \textbf{Drug} & \textbf{All} & \textbf{Diag.} & \textbf{Drug} & \textbf{All} \\
        \hline
        1 & 17.3 & 5.0 & 8.8 & 17.5 & 5.8 & 12.4 & 17.3  & 5.7 & 11.3 \\
        2 & 16.0 & 6.2 & 9.3 & 16.4 & 6.4 & 12.0 & 16.3 & 6.3 & 11.1 \\
        3 & 15.0 & 8.3 & 10.5 & 16.4 & 8.0 & 12.7 & 16.4 & 7.6 & 11.9 \\
        4 & 13.1 & 5.6 & 8.1 & 15.1 & 5.8 & 11.1 & 15.0 & 6.1 & 10.4 \\
        \textbf{5} & \textbf{38.4} & \textbf{74.7} & \textbf{63.0} & \textbf{34.4} & \textbf{73.8} & \textbf{51.5} & \textbf{34.8} & \textbf{55.1} & \textbf{55.1} \\
        \hline
    \end{tabular}
    \vspace{-0.4cm}
\end{table}

\subsubsection{Case Study} As a case study, we randomly selected a patient with a true survival time of $T=6.86$. Originally, CoxCC model predicted a survival time of $\hat{T}=6.032$ for this patient. Performing SurvAttack, the predicted survival time reduced to $\hat{T}^{*}=4.318$ (faking the patient as more urgent compared to other critical patients), while still maintaining a 0.96\% medical semantic similarity with the original intact subject. Table \ref{tab:case-study} presents the performance details of SurvAttack on this particular subject. SurvAttack employed a range of adversarial actions to achieve its objectives. For instance, as for replacement, SurvAttack substituted the drug code N05CA with N05CD in visit 5. These two codes fall under the N05C group, with co-occurrence of $P(C09DA|C09DX)>=0.75$. Both of these drugs belong to the category of "N05C: Hypnotics and Sedatives" and are commonly used for various medical treatments, including reducing anxiety and promoting relaxation.


%% file: tex/related_work.tex
\section{Related Work}

We review recent advancements in white-box and black-box adversarial attacks, highlighting algorithms that enhance our understanding of deep network robustness. Finally, we examine healthcare-based attacks and distinguish our method from them.

In \textbf{white-box} adversarial attacks, where accessing the structure and parameters of the model is plausible, most of the methods heavily depend on gradient calculation with respect to input features. Fast Gradient Sign Method (FGSM) introduced in \cite{goodfellow2014explaining} leverages gradients from the loss function concerning the input data to identify the perturbation direction that maximizes the loss subject to an $L_{\infty}$ constraint, causing incorrect model predictions \cite{kurakin2016adversarial,kurakin2016adversaria}. 
Jacobian-based Saliency Map Algorithm (JSMA) \cite{papernot2016limitations} iteratively computes a saliency mapping between inputs and outputs using the Jacobian matrix, to identify the most influential pixel (input feature) to target for perturbation. The Carlini and Wagner (C\&W)  \cite{carlini2017towards} attack is a sophisticated optimization-based attack. Focusing on the logit layer of a neural network, it identifies minimal input perturbations while keeping the perturbed input close to the original (Euclidean distance). Researchers have applied the above white-box techniques to other domains like natural language processing \cite{ebrahimi2017hotflip, yuan2021bridge, papernot2016crafting, zang2019word}. For example, LeapAttack \cite{ye2022leapattack}, a gradient-based method for generating high-quality text adversarial examples in the hard-label setting, optimizes perturbations using word embedding spaces to achieve high semantic similarity and low perturbation rates. However, all of these methods, although theoretically valuable, are often impractical and unrealistic as they require complete knowledge of and access to the target model's architecture, parameters, and training data, which is rarely accessible. Furthermore, these attacks require gradient calculations, making them computationally intensive and time-consuming, and thus impractical for large-scale use.

 The \textbf{black-box} scenario is a more realistic setting for the attack, where the victim model is inaccessible. A wide range of black-box techniques based on greedy-based scoring strategies \cite{ren2019generating, li2018textbugger, samanta2017towards, gao2018black}, reinforcement learning \cite{wei2022simultaneously,ye2022medattacker}, genetic algorithms \cite{chen2019poba, maheshwary2021generating}, etc., have been developed to tackle this challenge.  \cite{samanta2017towards} only uses word saliency to perform three modifications of removal, adding, and replacement to attack text classification. Probability weighted word saliency (PWWS) \cite{ren2019generating} introduces a black-box score-base adversarial attack method on text classification based on word replacement. TextBugger \cite{li2018textbugger} generates adversarial text by identifying the most critical sentences and the most significant words within those sentences and initiates the attack by replacing these important words using a bug-generation strategy. DeepWordBug \cite{gao2018black} creates subtle text perturbations in a black-box setting using novel sequential-aware scoring. This identifies key words for modification, causing the deep classifier to make incorrect predictions. Moreover, another black-box technique is transfer-based attacks which rely on training a substitute model with decision boundaries similar to the target model \cite{dong2019evading,fang2022learning, dong2021query, wei2022towards}. For instance, VQATTACK \cite{yin2024vqattack}, a transferable adversarial attack that generates adversarial samples using pre-trained models to target different black-box victim models, jointly updates image and text perturbations. 
 Transfer-based attacks are costly and quite unrealistic because they require access to training data, training a substitute model, and assume successful transfer of adversarial examples between different architectures.


Adversarial attack on EHR-based tasks is an emerging research area, especially in black-box scenarios. As a white-box attack method on EHRs, Longitudinal AdVersarial Attack (LAVA) \cite{an2019longitudinal} jointly models a saliency map and an attention mechanism, thereby identifying the optimal EHR feature at a specific visit for perturbation. \cite{sun2018identify} proposes an adversarial attack method for the continuous medical measurements in EHR data which solves a sparsity-regularized attack objective to target an LSTM model. 
MedAttacker \cite{ye2022medattacker}, a black-box adversarial attack method targeting EHR prediction models, addresses the unique challenges of EHRs through position selection and substitute selection by leveraging reinforcement learning. What distinguishes our study from prior works mainly are 1) SurvAttack is the first black-box adversarial attack method against ``patient survival models'', which, unlike traditional classification or regression tasks, primarily performs a survival ranking task to prioritize patients’ survival urgency. 
Second, as opposed to previous methods, our adversarial perturbations are clinically meaningful and compatible as we introduced an ontology-informed Synonym Code Selection (SCS) Strategy by leveraging domain knowledge and population-level co-occurrence statistics, and 3) we developed a deep embedding-based Semantic Similarity Function (SSF), evaluating the stealthiness of attack in latent space instead of relying on the naive count of perturbations.

\vspace{-0.2cm}

%% file: tex/Conclusion.tex
\section{Conclusion}

Ensuring the robustness of EHR-based patient survival models is vital due to their direct impact on human lives. To advance this field, we introduce SurvAttack, a black-box adversarial attack framework designed to compromise the robustness of survival models. SurvAttack is based on meticulously calculating a composite score for a set of adversarial actions while taking into account the clinical semantic similarity and survival output change, to diminish the survival ranking as well as the time regression predictive ability of patient survival models. Experimental results, including baseline comparisons, attack pattern analysis, survival explainability, and case studies, underscore the significance of our work.

%% file: Main.bbl

\begin{thebibliography}{50}


\ifx \showCODEN    \undefined \def \showCODEN     #1{\unskip}     \fi
\ifx \showDOI      \undefined \def \showDOI       #1{#1}\fi
\ifx \showISBNx    \undefined \def \showISBNx     #1{\unskip}     \fi
\ifx \showISBNxiii \undefined \def \showISBNxiii  #1{\unskip}     \fi
\ifx \showISSN     \undefined \def \showISSN      #1{\unskip}     \fi
\ifx \showLCCN     \undefined \def \showLCCN      #1{\unskip}     \fi
\ifx \shownote     \undefined \def \shownote      #1{#1}          \fi
\ifx \showarticletitle \undefined \def \showarticletitle #1{#1}   \fi
\ifx \showURL      \undefined \def \showURL       {\relax}        \fi
\providecommand\bibfield[2]{#2}
\providecommand\bibinfo[2]{#2}
\providecommand\natexlab[1]{#1}
\providecommand\showeprint[2][]{arXiv:#2}

\bibitem[Alge et~al\mbox{.}(2021)]%
        {alge2021two}
\bibfield{author}{\bibinfo{person}{Joseph Alge}, \bibinfo{person}{Kristin Dolan}, \bibinfo{person}{Joseph Angelo}, \bibinfo{person}{Sameer Thadani}, \bibinfo{person}{Manpreet Virk}, {and} \bibinfo{person}{Ayse Akcan~Arikan}.} \bibinfo{year}{2021}\natexlab{}.
\newblock \showarticletitle{Two to tango: kidney-lung interaction in acute kidney injury and acute respiratory distress syndrome}.
\newblock \bibinfo{journal}{\emph{Frontiers in Pediatrics}} (\bibinfo{year}{2021}), \bibinfo{pages}{1046}.
\newblock


\bibitem[An et~al\mbox{.}(2019)]%
        {an2019longitudinal}
\bibfield{author}{\bibinfo{person}{Sungtae An}, \bibinfo{person}{Cao Xiao}, \bibinfo{person}{Walter~F Stewart}, {and} \bibinfo{person}{Jimeng Sun}.} \bibinfo{year}{2019}\natexlab{}.
\newblock \showarticletitle{Longitudinal adversarial attack on electronic health records data}. In \bibinfo{booktitle}{\emph{The world wide web conference}}. \bibinfo{pages}{2558--2564}.
\newblock


\bibitem[Carlini and Wagner(2017)]%
        {carlini2017towards}
\bibfield{author}{\bibinfo{person}{Nicholas Carlini} {and} \bibinfo{person}{David Wagner}.} \bibinfo{year}{2017}\natexlab{}.
\newblock \showarticletitle{Towards evaluating the robustness of neural networks}. In \bibinfo{booktitle}{\emph{2017 ieee symposium on security and privacy (sp)}}. Ieee, \bibinfo{pages}{39--57}.
\newblock


\bibitem[Chen et~al\mbox{.}(2019)]%
        {chen2019poba}
\bibfield{author}{\bibinfo{person}{Jinyin Chen}, \bibinfo{person}{Mengmeng Su}, \bibinfo{person}{Shijing Shen}, \bibinfo{person}{Hui Xiong}, {and} \bibinfo{person}{Haibin Zheng}.} \bibinfo{year}{2019}\natexlab{}.
\newblock \showarticletitle{POBA-GA: Perturbation optimized black-box adversarial attacks via genetic algorithm}.
\newblock \bibinfo{journal}{\emph{Computers \& Security}}  \bibinfo{volume}{85} (\bibinfo{year}{2019}), \bibinfo{pages}{89--106}.
\newblock


\bibitem[Choi et~al\mbox{.}(2017)]%
        {choi2017gram}
\bibfield{author}{\bibinfo{person}{Edward Choi}, \bibinfo{person}{Mohammad~Taha Bahadori}, \bibinfo{person}{Le Song}, \bibinfo{person}{Walter~F Stewart}, {and} \bibinfo{person}{Jimeng Sun}.} \bibinfo{year}{2017}\natexlab{}.
\newblock \showarticletitle{GRAM: graph-based attention model for healthcare representation learning}. In \bibinfo{booktitle}{\emph{Proceedings of the 23rd ACM SIGKDD international conference on knowledge discovery and data mining}}. \bibinfo{pages}{787--795}.
\newblock


\bibitem[Choi et~al\mbox{.}(2016)]%
        {choi2016retain}
\bibfield{author}{\bibinfo{person}{Edward Choi}, \bibinfo{person}{Mohammad~Taha Bahadori}, \bibinfo{person}{Jimeng Sun}, \bibinfo{person}{Joshua Kulas}, \bibinfo{person}{Andy Schuetz}, {and} \bibinfo{person}{Walter Stewart}.} \bibinfo{year}{2016}\natexlab{}.
\newblock \showarticletitle{Retain: An interpretable predictive model for healthcare using reverse time attention mechanism}.
\newblock \bibinfo{journal}{\emph{Advances in neural information processing systems}}  \bibinfo{volume}{29} (\bibinfo{year}{2016}).
\newblock


\bibitem[Cox(1972)]%
        {cox1972regression}
\bibfield{author}{\bibinfo{person}{David~R Cox}.} \bibinfo{year}{1972}\natexlab{}.
\newblock \showarticletitle{Regression models and life-tables}.
\newblock \bibinfo{journal}{\emph{Journal of the Royal Statistical Society: Series B (Methodological)}} \bibinfo{volume}{34}, \bibinfo{number}{2} (\bibinfo{year}{1972}), \bibinfo{pages}{187--202}.
\newblock


\bibitem[Dong et~al\mbox{.}(2021)]%
        {dong2021query}
\bibfield{author}{\bibinfo{person}{Yinpeng Dong}, \bibinfo{person}{Shuyu Cheng}, \bibinfo{person}{Tianyu Pang}, \bibinfo{person}{Hang Su}, {and} \bibinfo{person}{Jun Zhu}.} \bibinfo{year}{2021}\natexlab{}.
\newblock \showarticletitle{Query-efficient black-box adversarial attacks guided by a transfer-based prior}.
\newblock \bibinfo{journal}{\emph{IEEE Transactions on Pattern Analysis and Machine Intelligence}} \bibinfo{volume}{44}, \bibinfo{number}{12} (\bibinfo{year}{2021}), \bibinfo{pages}{9536--9548}.
\newblock


\bibitem[Dong et~al\mbox{.}(2019)]%
        {dong2019evading}
\bibfield{author}{\bibinfo{person}{Yinpeng Dong}, \bibinfo{person}{Tianyu Pang}, \bibinfo{person}{Hang Su}, {and} \bibinfo{person}{Jun Zhu}.} \bibinfo{year}{2019}\natexlab{}.
\newblock \showarticletitle{Evading defenses to transferable adversarial examples by translation-invariant attacks}. In \bibinfo{booktitle}{\emph{Proceedings of the IEEE/CVF conference on computer vision and pattern recognition}}. \bibinfo{pages}{4312--4321}.
\newblock


\bibitem[Ebrahimi et~al\mbox{.}(2017)]%
        {ebrahimi2017hotflip}
\bibfield{author}{\bibinfo{person}{Javid Ebrahimi}, \bibinfo{person}{Anyi Rao}, \bibinfo{person}{Daniel Lowd}, {and} \bibinfo{person}{Dejing Dou}.} \bibinfo{year}{2017}\natexlab{}.
\newblock \showarticletitle{Hotflip: White-box adversarial examples for text classification}.
\newblock \bibinfo{journal}{\emph{arXiv preprint arXiv:1712.06751}} (\bibinfo{year}{2017}).
\newblock


\bibitem[Fang et~al\mbox{.}(2022)]%
        {fang2022learning}
\bibfield{author}{\bibinfo{person}{Shuman Fang}, \bibinfo{person}{Jie Li}, \bibinfo{person}{Xianming Lin}, {and} \bibinfo{person}{Rongrong Ji}.} \bibinfo{year}{2022}\natexlab{}.
\newblock \showarticletitle{Learning to learn transferable attack}. In \bibinfo{booktitle}{\emph{Proceedings of the AAAI Conference on Artificial Intelligence}}, Vol.~\bibinfo{volume}{36}. \bibinfo{pages}{571--579}.
\newblock


\bibitem[Faubel and Edelstein(2016)]%
        {faubel2016mechanisms}
\bibfield{author}{\bibinfo{person}{Sarah Faubel} {and} \bibinfo{person}{Charles~L Edelstein}.} \bibinfo{year}{2016}\natexlab{}.
\newblock \showarticletitle{Mechanisms and mediators of lung injury after acute kidney injury}.
\newblock \bibinfo{journal}{\emph{Nature Reviews Nephrology}} \bibinfo{volume}{12}, \bibinfo{number}{1} (\bibinfo{year}{2016}), \bibinfo{pages}{48--60}.
\newblock


\bibitem[Fotso(2018)]%
        {fotso2018deep}
\bibfield{author}{\bibinfo{person}{Stephane Fotso}.} \bibinfo{year}{2018}\natexlab{}.
\newblock \showarticletitle{Deep neural networks for survival analysis based on a multi-task framework}.
\newblock \bibinfo{journal}{\emph{arXiv preprint arXiv:1801.05512}} (\bibinfo{year}{2018}).
\newblock


\bibitem[Gao et~al\mbox{.}(2018)]%
        {gao2018black}
\bibfield{author}{\bibinfo{person}{Ji Gao}, \bibinfo{person}{Jack Lanchantin}, \bibinfo{person}{Mary~Lou Soffa}, {and} \bibinfo{person}{Yanjun Qi}.} \bibinfo{year}{2018}\natexlab{}.
\newblock \showarticletitle{Black-box generation of adversarial text sequences to evade deep learning classifiers}. In \bibinfo{booktitle}{\emph{2018 IEEE Security and Privacy Workshops (SPW)}}. IEEE, \bibinfo{pages}{50--56}.
\newblock


\bibitem[Goldblum et~al\mbox{.}(2022)]%
        {goldblum2022dataset}
\bibfield{author}{\bibinfo{person}{Micah Goldblum}, \bibinfo{person}{Dimitris Tsipras}, \bibinfo{person}{Chulin Xie}, \bibinfo{person}{Xinyun Chen}, \bibinfo{person}{Avi Schwarzschild}, \bibinfo{person}{Dawn Song}, \bibinfo{person}{Aleksander M{\k{a}}dry}, \bibinfo{person}{Bo Li}, {and} \bibinfo{person}{Tom Goldstein}.} \bibinfo{year}{2022}\natexlab{}.
\newblock \showarticletitle{Dataset security for machine learning: Data poisoning, backdoor attacks, and defenses}.
\newblock \bibinfo{journal}{\emph{IEEE Transactions on Pattern Analysis and Machine Intelligence}} \bibinfo{volume}{45}, \bibinfo{number}{2} (\bibinfo{year}{2022}), \bibinfo{pages}{1563--1580}.
\newblock


\bibitem[Goodfellow et~al\mbox{.}(2014)]%
        {goodfellow2014explaining}
\bibfield{author}{\bibinfo{person}{Ian~J Goodfellow}, \bibinfo{person}{Jonathon Shlens}, {and} \bibinfo{person}{Christian Szegedy}.} \bibinfo{year}{2014}\natexlab{}.
\newblock \showarticletitle{Explaining and harnessing adversarial examples}.
\newblock \bibinfo{journal}{\emph{arXiv preprint arXiv:1412.6572}} (\bibinfo{year}{2014}).
\newblock


\bibitem[Hadizadeh~Moghaddam et~al\mbox{.}(2024b)]%
        {10.1145/3709143}
\bibfield{author}{\bibinfo{person}{Arya Hadizadeh~Moghaddam}, \bibinfo{person}{Mohsen Nayebi~Kerdabadi}, \bibinfo{person}{Bin Liu}, \bibinfo{person}{Mei Liu}, {and} \bibinfo{person}{Zijun Yao}.} \bibinfo{year}{2024}\natexlab{b}.
\newblock \showarticletitle{Discovering Time-Aware Hidden Dependencies with Personalized Graphical Structure in Electronic Health Records}.
\newblock \bibinfo{journal}{\emph{ACM Trans. Knowl. Discov. Data}} (\bibinfo{date}{Dec.} \bibinfo{year}{2024}).
\newblock
\showISSN{1556-4681}
\urldef\tempurl%
\url{https://doi.org/10.1145/3709143}
\showDOI{\tempurl}
\newblock
\shownote{Just Accepted}.


\bibitem[Hadizadeh~Moghaddam et~al\mbox{.}(2024a)]%
        {hadizadeh2024contrastive}
\bibfield{author}{\bibinfo{person}{Arya Hadizadeh~Moghaddam}, \bibinfo{person}{Mohsen Nayebi~Kerdabadi}, \bibinfo{person}{Mei Liu}, {and} \bibinfo{person}{Zijun Yao}.} \bibinfo{year}{2024}\natexlab{a}.
\newblock \showarticletitle{Contrastive learning on medical intents for sequential prescription recommendation}. In \bibinfo{booktitle}{\emph{Proceedings of the 33rd ACM International Conference on Information and Knowledge Management}}. \bibinfo{pages}{748--757}.
\newblock


\bibitem[Harrell et~al\mbox{.}(1982)]%
        {harrell1982evaluating}
\bibfield{author}{\bibinfo{person}{Frank~E Harrell}, \bibinfo{person}{Robert~M Califf}, \bibinfo{person}{David~B Pryor}, \bibinfo{person}{Kerry~L Lee}, {and} \bibinfo{person}{Robert~A Rosati}.} \bibinfo{year}{1982}\natexlab{}.
\newblock \showarticletitle{Evaluating the yield of medical tests}.
\newblock \bibinfo{journal}{\emph{Jama}} \bibinfo{volume}{247}, \bibinfo{number}{18} (\bibinfo{year}{1982}), \bibinfo{pages}{2543--2546}.
\newblock


\bibitem[Kanic et~al\mbox{.}(2019)]%
        {kanic2019acute}
\bibfield{author}{\bibinfo{person}{Vojko Kanic}, \bibinfo{person}{Gregor Kompara}, \bibinfo{person}{David {\v{S}}uran}, \bibinfo{person}{Alojz Tapajner}, \bibinfo{person}{Franjo~Husam Naji}, {and} \bibinfo{person}{Andreja Sinkovic}.} \bibinfo{year}{2019}\natexlab{}.
\newblock \showarticletitle{Acute kidney injury in patients with myocardial infarction undergoing percutaneous coronary intervention using radial versus femoral access}.
\newblock \bibinfo{journal}{\emph{BMC nephrology}}  \bibinfo{volume}{20} (\bibinfo{year}{2019}), \bibinfo{pages}{1--7}.
\newblock


\bibitem[Katzman et~al\mbox{.}(2018)]%
        {katzman2018deepsurv}
\bibfield{author}{\bibinfo{person}{Jared~L Katzman}, \bibinfo{person}{Uri Shaham}, \bibinfo{person}{Alexander Cloninger}, \bibinfo{person}{Jonathan Bates}, \bibinfo{person}{Tingting Jiang}, {and} \bibinfo{person}{Yuval Kluger}.} \bibinfo{year}{2018}\natexlab{}.
\newblock \showarticletitle{DeepSurv: personalized treatment recommender system using a Cox proportional hazards deep neural network}.
\newblock \bibinfo{journal}{\emph{BMC medical research methodology}} \bibinfo{volume}{18}, \bibinfo{number}{1} (\bibinfo{year}{2018}), \bibinfo{pages}{1--12}.
\newblock


\bibitem[Kurakin et~al\mbox{.}(2016a)]%
        {kurakin2016adversaria}
\bibfield{author}{\bibinfo{person}{Alexey Kurakin}, \bibinfo{person}{Ian Goodfellow}, {and} \bibinfo{person}{Samy Bengio}.} \bibinfo{year}{2016}\natexlab{a}.
\newblock \showarticletitle{Adversarial machine learning at scale}.
\newblock \bibinfo{journal}{\emph{arXiv preprint arXiv:1611.01236}} (\bibinfo{year}{2016}).
\newblock


\bibitem[Kurakin et~al\mbox{.}(2016b)]%
        {kurakin2016adversarial}
\bibfield{author}{\bibinfo{person}{Alexey Kurakin}, \bibinfo{person}{Ian Goodfellow}, \bibinfo{person}{Samy Bengio}, {et~al\mbox{.}}} \bibinfo{year}{2016}\natexlab{b}.
\newblock \bibinfo{title}{Adversarial examples in the physical world}.
\newblock
\newblock


\bibitem[Kvamme et~al\mbox{.}(2019)]%
        {JMLR:v20:18-424}
\bibfield{author}{\bibinfo{person}{H{{\aa}}vard Kvamme}, \bibinfo{person}{{{\O}}rnulf Borgan}, {and} \bibinfo{person}{Ida Scheel}.} \bibinfo{year}{2019}\natexlab{}.
\newblock \showarticletitle{Time-to-Event Prediction with Neural Networks and Cox Regression}.
\newblock \bibinfo{journal}{\emph{Journal of Machine Learning Research}} \bibinfo{volume}{20}, \bibinfo{number}{129} (\bibinfo{year}{2019}), \bibinfo{pages}{1--30}.
\newblock
\urldef\tempurl%
\url{http://jmlr.org/papers/v20/18-424.html}
\showURL{%
\tempurl}


\bibitem[Lee et~al\mbox{.}(2018)]%
        {lee2018deephit}
\bibfield{author}{\bibinfo{person}{Changhee Lee}, \bibinfo{person}{William Zame}, \bibinfo{person}{Jinsung Yoon}, {and} \bibinfo{person}{Mihaela Van Der~Schaar}.} \bibinfo{year}{2018}\natexlab{}.
\newblock \showarticletitle{Deephit: A deep learning approach to survival analysis with competing risks}. In \bibinfo{booktitle}{\emph{Proceedings of the AAAI conference on artificial intelligence}}, Vol.~\bibinfo{volume}{32}.
\newblock


\bibitem[Li et~al\mbox{.}(2018)]%
        {li2018textbugger}
\bibfield{author}{\bibinfo{person}{Jinfeng Li}, \bibinfo{person}{Shouling Ji}, \bibinfo{person}{Tianyu Du}, \bibinfo{person}{Bo Li}, {and} \bibinfo{person}{Ting Wang}.} \bibinfo{year}{2018}\natexlab{}.
\newblock \showarticletitle{Textbugger: Generating adversarial text against real-world applications}.
\newblock \bibinfo{journal}{\emph{arXiv preprint arXiv:1812.05271}} (\bibinfo{year}{2018}).
\newblock


\bibitem[Liu et~al\mbox{.}(2018)]%
        {liu2018early}
\bibfield{author}{\bibinfo{person}{Bin Liu}, \bibinfo{person}{Ying Li}, \bibinfo{person}{Zhaonan Sun}, \bibinfo{person}{Soumya Ghosh}, {and} \bibinfo{person}{Kenney Ng}.} \bibinfo{year}{2018}\natexlab{}.
\newblock \showarticletitle{Early prediction of diabetes complications from electronic health records: A multi-task survival analysis approach}. In \bibinfo{booktitle}{\emph{Proceedings of the AAAI Conference on Artificial Intelligence}}, Vol.~\bibinfo{volume}{32}.
\newblock


\bibitem[Liu et~al\mbox{.}(2022)]%
        {liu2022development}
\bibfield{author}{\bibinfo{person}{Kang Liu}, \bibinfo{person}{Xiangzhou Zhang}, \bibinfo{person}{Weiqi Chen}, \bibinfo{person}{SL Alan}, \bibinfo{person}{John~A Kellum}, \bibinfo{person}{Michael~E Matheny}, \bibinfo{person}{Steven~Q Simpson}, \bibinfo{person}{Yong Hu}, {and} \bibinfo{person}{Mei Liu}.} \bibinfo{year}{2022}\natexlab{}.
\newblock \showarticletitle{Development and validation of a personalized model with transfer learning for acute kidney injury risk estimation using electronic health records}.
\newblock \bibinfo{journal}{\emph{JAMA Network Open}} \bibinfo{volume}{5}, \bibinfo{number}{7} (\bibinfo{year}{2022}), \bibinfo{pages}{e2219776--e2219776}.
\newblock


\bibitem[Maheshwary et~al\mbox{.}(2021)]%
        {maheshwary2021generating}
\bibfield{author}{\bibinfo{person}{Rishabh Maheshwary}, \bibinfo{person}{Saket Maheshwary}, {and} \bibinfo{person}{Vikram Pudi}.} \bibinfo{year}{2021}\natexlab{}.
\newblock \showarticletitle{Generating natural language attacks in a hard label black box setting}. In \bibinfo{booktitle}{\emph{Proceedings of the AAAI Conference on Artificial Intelligence}}, Vol.~\bibinfo{volume}{35}. \bibinfo{pages}{13525--13533}.
\newblock


\bibitem[Moghaddam et~al\mbox{.}(2024)]%
        {moghaddam2024meta}
\bibfield{author}{\bibinfo{person}{Arya~Hadizadeh Moghaddam}, \bibinfo{person}{Mohsen~Nayebi Kerdabadi}, \bibinfo{person}{Cuncong Zhong}, {and} \bibinfo{person}{Zijun Yao}.} \bibinfo{year}{2024}\natexlab{}.
\newblock \showarticletitle{Meta-Learning on Augmented Gene Expression Profiles for Enhanced Lung Cancer Detection}.
\newblock \bibinfo{journal}{\emph{arXiv preprint arXiv:2408.09635}} (\bibinfo{year}{2024}).
\newblock


\bibitem[Moosavi-Dezfooli et~al\mbox{.}(2016)]%
        {moosavi2016deepfool}
\bibfield{author}{\bibinfo{person}{Seyed-Mohsen Moosavi-Dezfooli}, \bibinfo{person}{Alhussein Fawzi}, {and} \bibinfo{person}{Pascal Frossard}.} \bibinfo{year}{2016}\natexlab{}.
\newblock \showarticletitle{Deepfool: a simple and accurate method to fool deep neural networks}. In \bibinfo{booktitle}{\emph{Proceedings of the IEEE conference on computer vision and pattern recognition}}. \bibinfo{pages}{2574--2582}.
\newblock


\bibitem[Nayebi~Kerdabadi et~al\mbox{.}(2023)]%
        {nayebi2023contrastive}
\bibfield{author}{\bibinfo{person}{Mohsen Nayebi~Kerdabadi}, \bibinfo{person}{Arya Hadizadeh~Moghaddam}, \bibinfo{person}{Bin Liu}, \bibinfo{person}{Mei Liu}, {and} \bibinfo{person}{Zijun Yao}.} \bibinfo{year}{2023}\natexlab{}.
\newblock \showarticletitle{Contrastive learning of temporal distinctiveness for survival analysis in electronic health records}. In \bibinfo{booktitle}{\emph{Proceedings of the 32nd ACM International Conference on Information and Knowledge Management}}. \bibinfo{pages}{1897--1906}.
\newblock


\bibitem[Ohno-Machado(2001)]%
        {ohno2001modeling}
\bibfield{author}{\bibinfo{person}{Lucila Ohno-Machado}.} \bibinfo{year}{2001}\natexlab{}.
\newblock \showarticletitle{Modeling medical prognosis: survival analysis techniques}.
\newblock \bibinfo{journal}{\emph{Journal of biomedical informatics}} \bibinfo{volume}{34}, \bibinfo{number}{6} (\bibinfo{year}{2001}), \bibinfo{pages}{428--439}.
\newblock


\bibitem[Papernot et~al\mbox{.}(2016a)]%
        {papernot2016limitations}
\bibfield{author}{\bibinfo{person}{Nicolas Papernot}, \bibinfo{person}{Patrick McDaniel}, \bibinfo{person}{Somesh Jha}, \bibinfo{person}{Matt Fredrikson}, \bibinfo{person}{Z~Berkay Celik}, {and} \bibinfo{person}{Ananthram Swami}.} \bibinfo{year}{2016}\natexlab{a}.
\newblock \showarticletitle{The limitations of deep learning in adversarial settings}. In \bibinfo{booktitle}{\emph{2016 IEEE European symposium on security and privacy (EuroS\&P)}}. IEEE, \bibinfo{pages}{372--387}.
\newblock


\bibitem[Papernot et~al\mbox{.}(2016b)]%
        {papernot2016crafting}
\bibfield{author}{\bibinfo{person}{Nicolas Papernot}, \bibinfo{person}{Patrick McDaniel}, \bibinfo{person}{Ananthram Swami}, {and} \bibinfo{person}{Richard Harang}.} \bibinfo{year}{2016}\natexlab{b}.
\newblock \showarticletitle{Crafting adversarial input sequences for recurrent neural networks}. In \bibinfo{booktitle}{\emph{MILCOM 2016-2016 IEEE Military Communications Conference}}. IEEE, \bibinfo{pages}{49--54}.
\newblock


\bibitem[Park and Faubel(2021)]%
        {park2021acute}
\bibfield{author}{\bibinfo{person}{Bryan~D Park} {and} \bibinfo{person}{Sarah Faubel}.} \bibinfo{year}{2021}\natexlab{}.
\newblock \showarticletitle{Acute kidney injury and acute respiratory distress syndrome}.
\newblock \bibinfo{journal}{\emph{Critical Care Clinics}} \bibinfo{volume}{37}, \bibinfo{number}{4} (\bibinfo{year}{2021}), \bibinfo{pages}{835--849}.
\newblock


\bibitem[Ren et~al\mbox{.}(2019)]%
        {ren2019generating}
\bibfield{author}{\bibinfo{person}{Shuhuai Ren}, \bibinfo{person}{Yihe Deng}, \bibinfo{person}{Kun He}, {and} \bibinfo{person}{Wanxiang Che}.} \bibinfo{year}{2019}\natexlab{}.
\newblock \showarticletitle{Generating natural language adversarial examples through probability weighted word saliency}. In \bibinfo{booktitle}{\emph{Proceedings of the 57th annual meeting of the association for computational linguistics}}. \bibinfo{pages}{1085--1097}.
\newblock


\bibitem[Samanta and Mehta(2017)]%
        {samanta2017towards}
\bibfield{author}{\bibinfo{person}{Suranjana Samanta} {and} \bibinfo{person}{Sameep Mehta}.} \bibinfo{year}{2017}\natexlab{}.
\newblock \showarticletitle{Towards crafting text adversarial samples}.
\newblock \bibinfo{journal}{\emph{arXiv preprint arXiv:1707.02812}} (\bibinfo{year}{2017}).
\newblock


\bibitem[Shacham et~al\mbox{.}(2016)]%
        {shacham2016acute}
\bibfield{author}{\bibinfo{person}{Yacov Shacham}, \bibinfo{person}{Arie Steinvil}, {and} \bibinfo{person}{Yaron Arbel}.} \bibinfo{year}{2016}\natexlab{}.
\newblock \showarticletitle{Acute kidney injury among ST elevation myocardial infarction patients treated by primary percutaneous coronary intervention: a multifactorial entity}.
\newblock \bibinfo{journal}{\emph{Journal of nephrology}}  \bibinfo{volume}{29} (\bibinfo{year}{2016}), \bibinfo{pages}{169--174}.
\newblock


\bibitem[Sun et~al\mbox{.}(2018)]%
        {sun2018identify}
\bibfield{author}{\bibinfo{person}{Mengying Sun}, \bibinfo{person}{Fengyi Tang}, \bibinfo{person}{Jinfeng Yi}, \bibinfo{person}{Fei Wang}, {and} \bibinfo{person}{Jiayu Zhou}.} \bibinfo{year}{2018}\natexlab{}.
\newblock \showarticletitle{Identify susceptible locations in medical records via adversarial attacks on deep predictive models}. In \bibinfo{booktitle}{\emph{Proceedings of the 24th ACM SIGKDD international conference on knowledge discovery \& data mining}}. \bibinfo{pages}{793--801}.
\newblock


\bibitem[Vaswani et~al\mbox{.}(2017)]%
        {vaswani2017attention}
\bibfield{author}{\bibinfo{person}{Ashish Vaswani}, \bibinfo{person}{Noam Shazeer}, \bibinfo{person}{Niki Parmar}, \bibinfo{person}{Jakob Uszkoreit}, \bibinfo{person}{Llion Jones}, \bibinfo{person}{Aidan~N Gomez}, \bibinfo{person}{{\L}ukasz Kaiser}, {and} \bibinfo{person}{Illia Polosukhin}.} \bibinfo{year}{2017}\natexlab{}.
\newblock \showarticletitle{Attention is all you need}.
\newblock \bibinfo{journal}{\emph{Advances in neural information processing systems}}  \bibinfo{volume}{30} (\bibinfo{year}{2017}).
\newblock


\bibitem[Wang et~al\mbox{.}(2019)]%
        {wang2019machine}
\bibfield{author}{\bibinfo{person}{Ping Wang}, \bibinfo{person}{Yan Li}, {and} \bibinfo{person}{Chandan~K Reddy}.} \bibinfo{year}{2019}\natexlab{}.
\newblock \showarticletitle{Machine learning for survival analysis: A survey}.
\newblock \bibinfo{journal}{\emph{ACM Computing Surveys (CSUR)}} \bibinfo{volume}{51}, \bibinfo{number}{6} (\bibinfo{year}{2019}), \bibinfo{pages}{1--36}.
\newblock


\bibitem[Wei et~al\mbox{.}(2022b)]%
        {wei2022simultaneously}
\bibfield{author}{\bibinfo{person}{Xingxing Wei}, \bibinfo{person}{Ying Guo}, \bibinfo{person}{Jie Yu}, {and} \bibinfo{person}{Bo Zhang}.} \bibinfo{year}{2022}\natexlab{b}.
\newblock \showarticletitle{Simultaneously optimizing perturbations and positions for black-box adversarial patch attacks}.
\newblock \bibinfo{journal}{\emph{IEEE transactions on pattern analysis and machine intelligence}} (\bibinfo{year}{2022}).
\newblock


\bibitem[Wei et~al\mbox{.}(2022a)]%
        {wei2022towards}
\bibfield{author}{\bibinfo{person}{Zhipeng Wei}, \bibinfo{person}{Jingjing Chen}, \bibinfo{person}{Micah Goldblum}, \bibinfo{person}{Zuxuan Wu}, \bibinfo{person}{Tom Goldstein}, {and} \bibinfo{person}{Yu-Gang Jiang}.} \bibinfo{year}{2022}\natexlab{a}.
\newblock \showarticletitle{Towards transferable adversarial attacks on vision transformers}. In \bibinfo{booktitle}{\emph{Proceedings of the AAAI Conference on Artificial Intelligence}}, Vol.~\bibinfo{volume}{36}. \bibinfo{pages}{2668--2676}.
\newblock


\bibitem[Ye et~al\mbox{.}(2022a)]%
        {ye2022leapattack}
\bibfield{author}{\bibinfo{person}{Muchao Ye}, \bibinfo{person}{Jinghui Chen}, \bibinfo{person}{Chenglin Miao}, \bibinfo{person}{Ting Wang}, {and} \bibinfo{person}{Fenglong Ma}.} \bibinfo{year}{2022}\natexlab{a}.
\newblock \showarticletitle{Leapattack: Hard-label adversarial attack on text via gradient-based optimization}. In \bibinfo{booktitle}{\emph{Proceedings of the 28th ACM SIGKDD Conference on Knowledge Discovery and Data Mining}}. \bibinfo{pages}{2307--2315}.
\newblock


\bibitem[Ye et~al\mbox{.}(2022b)]%
        {ye2022medattacker}
\bibfield{author}{\bibinfo{person}{Muchao Ye}, \bibinfo{person}{Junyu Luo}, \bibinfo{person}{Guanjie Zheng}, \bibinfo{person}{Cao Xiao}, \bibinfo{person}{Houping Xiao}, \bibinfo{person}{Ting Wang}, {and} \bibinfo{person}{Fenglong Ma}.} \bibinfo{year}{2022}\natexlab{b}.
\newblock \showarticletitle{MedAttacker: Exploring black-box adversarial attacks on risk prediction models in healthcare}. In \bibinfo{booktitle}{\emph{2022 IEEE International Conference on Bioinformatics and Biomedicine (BIBM)}}. IEEE, \bibinfo{pages}{1777--1780}.
\newblock


\bibitem[Yin et~al\mbox{.}(2024)]%
        {yin2024vqattack}
\bibfield{author}{\bibinfo{person}{Ziyi Yin}, \bibinfo{person}{Muchao Ye}, \bibinfo{person}{Tianrong Zhang}, \bibinfo{person}{Jiaqi Wang}, \bibinfo{person}{Han Liu}, \bibinfo{person}{Jinghui Chen}, \bibinfo{person}{Ting Wang}, {and} \bibinfo{person}{Fenglong Ma}.} \bibinfo{year}{2024}\natexlab{}.
\newblock \showarticletitle{VQAttack: Transferable Adversarial Attacks on Visual Question Answering via Pre-trained Models}.
\newblock \bibinfo{journal}{\emph{arXiv preprint arXiv:2402.11083}} (\bibinfo{year}{2024}).
\newblock


\bibitem[Yu et~al\mbox{.}(2011)]%
        {yu2011learning}
\bibfield{author}{\bibinfo{person}{Chun-Nam Yu}, \bibinfo{person}{Russell Greiner}, \bibinfo{person}{Hsiu-Chin Lin}, {and} \bibinfo{person}{Vickie Baracos}.} \bibinfo{year}{2011}\natexlab{}.
\newblock \showarticletitle{Learning patient-specific cancer survival distributions as a sequence of dependent regressors}.
\newblock \bibinfo{journal}{\emph{Advances in neural information processing systems}}  \bibinfo{volume}{24} (\bibinfo{year}{2011}).
\newblock


\bibitem[Yuan et~al\mbox{.}(2021)]%
        {yuan2021bridge}
\bibfield{author}{\bibinfo{person}{Lifan Yuan}, \bibinfo{person}{Yichi Zhang}, \bibinfo{person}{Yangyi Chen}, {and} \bibinfo{person}{Wei Wei}.} \bibinfo{year}{2021}\natexlab{}.
\newblock \showarticletitle{Bridge the gap between cv and nlp! a gradient-based textual adversarial attack framework}.
\newblock \bibinfo{journal}{\emph{arXiv preprint arXiv:2110.15317}} (\bibinfo{year}{2021}).
\newblock


\bibitem[Zang et~al\mbox{.}(2019)]%
        {zang2019word}
\bibfield{author}{\bibinfo{person}{Yuan Zang}, \bibinfo{person}{Fanchao Qi}, \bibinfo{person}{Chenghao Yang}, \bibinfo{person}{Zhiyuan Liu}, \bibinfo{person}{Meng Zhang}, \bibinfo{person}{Qun Liu}, {and} \bibinfo{person}{Maosong Sun}.} \bibinfo{year}{2019}\natexlab{}.
\newblock \showarticletitle{Word-level textual adversarial attacking as combinatorial optimization}.
\newblock \bibinfo{journal}{\emph{arXiv preprint arXiv:1910.12196}} (\bibinfo{year}{2019}).
\newblock


\end{thebibliography}
